\documentclass[journal]{IEEEtran}
\usepackage{widetext}

\usepackage{booktabs}
\usepackage{multirow}
\usepackage{xcolor}
\usepackage[normalem]{ulem}

\ifCLASSINFOpdf
  \usepackage[pdftex]{graphicx}
  \graphicspath{{../pdf/}{../jpeg/}}
  \DeclareGraphicsExtensions{.pdf,.jpeg,.png}
\else
  \usepackage[dvips]{graphicx}
  \graphicspath{{../eps/}}
  \DeclareGraphicsExtensions{.eps}
\fi
\usepackage{xcolor}

\usepackage{amsmath}
\usepackage{amsfonts}
\usepackage{bm}

\usepackage{subcaption}

\begin{document}
%
\title{Image quality prediction using synthetic and natural codebooks: comparative results \\
\vspace {0.5cm}
\large{\bf Huawei technical report \\}
}

%
%
%

\author{Maxim Koroteev, Kirill Aistov, Valeriy Berezovskiy, and Pavel Frolov
\thanks{Moscow Research Center, Moscow, Russia. Cloud BU.  E-mail: koroteev.maxim@huawei.com}}

%
%

\markboth{Huawei Technical Report}
{Koroteev \MakeLowercase{\textit{et al.}}:}
%



\maketitle

\begin{abstract}
We investigate a model for image/video quality assessment based on building a set of
codevectors representing in a sense some basic properties of images, similar to well-known CORNIA model. We analyze the codebook building method and propose some modifications for it. Also the algorithm is investigated from the point of inference time reduction. Both natural and synthetic images are used for building codebooks and some analysis of synthetic images used for codebooks is provided. It is demonstrated the results on quality assessment may be improves with the use if synthetic images for codebook construction. We also demonstrate regimes of the algorithm in which real time execution on CPU is possible for sufficiently high correlations with mean opinion score (MOS). Various pooling strategies are considered as well as the problem of metric sensitivity to bitrate.
\end{abstract}


%
\IEEEpeerreviewmaketitle

\section{Introduction}
%
%
%
%

\IEEEPARstart{A}mong the applications of image processing image quality assessment occupies a noticeable place, especially in last decades when the need for better compression of images and video resulted in the necessity of quality estimation algorithms in real time. 

As it is well known, approaches for image quality assessment can be divided in several classes depending on whether the reference image is avaliable and to what extent it can be used. If for the full reference (FR) methods the reference image is used essentially, for non-reference (NR)\footnote{Strangely enough in the literature we see the use of the term `no-reference'. Not trying to argue with the community, we only note that normative English grammar does not know the prefix `no'; in similar situations prefix `non' should be used. So throughout this paper we use the term `non-reference', which is grammatically correct.} approaches only the image or the video frame which is processed at the moment is utilized for qulaity assessment without referring to some original image. 

On the other hand, methods for quality assessment can be divided according to the approaches applied for solving this problem. Here two main groups of approaches can be noticed: those using neural networks where features are designed automatically (or semi-autimatically) by the algorithm and those exploiting features engineered based on natural scene statistics (NSS). We in this paper are interested in the non-reference (NR) approach and in terms of the method we are specifically based on an earlier proposed algorithm called CORNIA\cite{CORNIA} which represents a kind of intermediate approach. Neural networks are not applied directly but the pipeline constructed in the method for solving the task of quality assessment may include hand-made features and in the same time partially resembles building a NN even though the learning procedure is used only as the last step for a regression algorithm.

A number of modifications have been proposed for the CORNIA algorithm. 
In \cite{SF100} authors introduce a supervised codebook (filter) learning method based on stochastic gradient descent instead of unsupervised learning via k-means.
In QAF model \cite{QAF} additional image features are extracted using Gabor filters; the codebook is created via sparse filter learning.
HOSA model \cite{HOSA} utilizes a feature encoding method in which a number of higher order statistics are calculated between image patches and cluster centers. 

One of the interesting observations, published quite recently \cite{Bosse}, was that the codebooks built using some random patches of pixels instead of patches extracted from real images can provide sufficiently good results for image quality. Based on this observation we exploited the possibility of using synthetic images as the building blocks for the quality assessment model. These synthetic images are random but their pixels are correlated. The correlations are characterized by the power-law decaying power spectrum. The models of that kind were investigated back some twenty five years ago\cite{Mumford} and still attract some attention even in applications to image quality assessment (see e.g., \cite{madhusudana})

In our approach we are primarily focused on creating an algorithm to be used in real-time applications, so we intorduce multiple modifications into the codebooks algorithm based on CORNIA to provide satisfactory results both for correlations and compexity (inference time). Specifically, among the most significant modifications we use the information obtained using both luminance and chrominance components of the image; it is demonstrated that this results in the improved correlations of the metric built with the mean opinion score (MOS). In addition, we investigate the application of synthetic codebooks, generated from synthetic images, some statistical properties of which being the same as for natural images.

We also apply the algorithm to the video data and analyze the application of two types of pooling (average pooling and std-pooling) which have some effect on the final correlations.

A specific part of this work is related to optimization of the current algorithms in terms of the model size and the inference time. 


\section{Some remarks on image processing in human visual system}
\label{sec:hvs}

Let us consider an image $I$ of the size $N\times N$ pixels. The fact that the image is assumed to be square is not too restrictive as the role of the number $N$ is to establish a characteristic length scale that can also be done in a certain way if the image is rectangular. 

The model of the image in the human eye assumes that the retina is represented as a lattice of cells, each of which processes a small patch of the image; the effective size of the patch to be processed is defined by the spread of the impulse response function of the cell \cite{wandell} (chps. 2, 5). To simulate this mechanism we have several options. One of them would be to divide $I$ into the grid of regular patches of size $n\times n$ pixels and process these patches. In this case the effective size of a patch $n$ is a parameter of the model and can be properly investigated. Alternatively, we can generalize this approach if we just sample a set of random patches from $I$ and use them to represent signals on the retina. It is clear that the latter approach has an advantage to be able to catch intersections of patches occurring for the neighbouring receptors of the retina because of the smoothness of the impulse response function, while the former implicitely assumes the impulse response function has a finite support. 

It is a well known fact that pixels of natural images are correlated, which results in the scale-invariant power-law spectra measured  in natural images\cite{RudermanBialek}. The analysis of how these correlations are processed in the cells on the retina was the subject of much research back in 90-s and 00-s\cite{Olshausen}. The evidence was found that the cells effectively remove correlations of closely located pixels, a process which can be thought of and simulated as whitening the signals in the collection of patches. Whitened signal are then transmitted further along the optical nerve to the brain.

These observations establish three main components of the model which aims at representing the process of human (and not only human) vision: 1) selection of a set of patches; 2) a procedure of whitening; 3) computations on the whitened signals. 

Another rationale for establishing this model from the point of view of image quality assessment is the use of mean opinion score (MOS) as an estimator, which implies that judgements of individuals about the quality of an image emerge as a result of a pre-processing of images by the human vision system. It is plausible to think that as MOS is built based on the main traits of the model above, then artificially designed metrics incorporating this model have a good chance to be highly correlated with MOS.

\section{The model and its properties}
\label{sec:model}

\subsection{Building the codebook}

Based in more or less extent on the ideas we reviewed in section \ref{sec:hvs}, which, however, should not be percieved too rigorously, a significant amount of models have been proposed for image quality assessment, which applied three main processes described above as their building blocks. One of the first and noticebly effective methods related to a codebook construction was CORNIA \cite{CORNIA} and our consideration here is based on main details of this model. We provide below the description of the model in the form how it was used in our analysis.

Let us consider a sample of $M$ images, each of them having the size $N\times N$ pixels. We would like to suggest a smaller set of generic {\it descriptors} for this image set, keeping in mind the descriptors should be coherent with the processing of the original set with the human visual system. So as a first step we sample random patches $\bar{x}_{i}$, $i=1,2\ldots m$ of the size $n\times n$ from each image, hence collecting $M\times m$ patches from our data set. These patches are then whitened together\footnote{there is a number of methods to be used for whitening; we numerically investigated some of them, see the next section.} and we obtain a set of processed patches for each image $\bar{y}_{i}$, $i=1,2,\ldots m$ which would represent the result of presenting the set to the human eye and processing in HVS (human visual system). For the patches under consideration we will use the name {\it vectors} below. Both $\bar{x}_{i}$ and $\bar{y}_{i}$ are elements of the space $\mathbb{R}^{d}$, where $d = n^{2}$. More formally, we create a matrix $X=[\bar{x}_{1}\; \bar{x}_{2}\ldots \bar{x}_{M\times m}]$, of the size $d\times (Mm)$ and apply a decorrelation transform $W$ to the columns of $X$ as follows (see \cite{Bethge} for a more detailed account). The correlation matrix of the data $X$ has the form $C = X\times X^{T}$. It can be decomposed as $C=UDU^{T}$, where $U$ is the matrix whose columns are the eigen vectors of $C$ and $D$ is a diagonal matrix which has eigen values of $C$ on its main diagonal. Then any second-order decorrelating transform has the form
\begin{equation}
\label{ZCA}
W = \tilde{D}\times V\times D^{-1/2}\times U^{T},
\end{equation}
where $U$ and $D$ were defined as above, $V$ -- is an arbitrary orthogonal matrix, and $\tilde{D}$ is an arbitrary diagonal matrix. Any particular choice of $V$ and $\tilde{D}$ will result in various decorrelating transforms \cite{Bethge}\footnote{It is worth to stress that the transforms we provided are based on the data matrix $X$. Another option could be an approximation of $U$ and $D$ by some standard basis, like Fourier. We will have a chance to discuss this in more detail below.}.

After applying $W$ we obtain the whitened data $Y=WX$. However the size of the matrix $Y$ is still $d\times (Mm)$. Its parameters are determined by the properties of the human visual system. Specifically, as we said above, $\sqrt{d}=n\sim$ a characteristic spread of the impulse response function; $m$ should be in some correspondence with the number of receptor cells on retina\footnote{probably in reality it is much lower.} and $M$ is the only parameter not related to the human vision process\footnote{Of course we are interested in relatively large $M$. Ideally we would like to collect information from each image in the data set; this guarantees that the model actually ``saw'' all the images. On the other hand, whether we need all images to be represented is determined by the resemblance of particular images to each other. An interesting problem would be to find a metric which would allow to estimate $M$ in a correct way in terms of image quality metrics.}. 

Assuming that after whitening and normalization many vectors may turn out to be similar we can reduce their number using a clusterization method, like e.g., K-means. Thus another parameter appearing in the model is $K$, the number of clusters; obviously it strongly affects the computation speed. Thus after clusterization we obtain a matrix $O$ of the size $d\times K$, which represents the data set and will be referred to as {\it a codebook} and its columns as {\it codevectors}.

\subsection{Correlating images on codevectors}
\label{subsec:correlating}

The codebook built, we assume that newly presented images, processed in the same way as codevectors can be compared to the vectors of the codebook in terms of some correlation measure. If $\tilde{Y}$ is the matrix whose columns correspond to the patches of a single image processed in the same way as the code vectors; let us assume we have $m$ such vectors as for the codevectors. We can measure how the codebook $O$ is correlated with $\tilde{Y}$ by computing the product $S = O^{T}\times\tilde{Y}$\footnote{Note we essentially compute multiple inner products in the eucledian space $R^{d}$ here. One can investigate more relevant norms for this model.}. The entries of matrix $S$ of the size $K\times m$ are inner products, which we denote as $s_{ij}$; we can further refine them by applying a non-linear soft-encoding function as follows. For each column $\bar{s}_{j}$ of $S$ let us build other two columns 
$$
\hat{s}^{+}_{j} = [\max(s_{1j},0)\; \max(s_{2j},0)\ldots \max(s_{Kj},0)]^{T},
$$
$$
\hat{s}^{-}_{j} = [\max(-s_{1j},0)\; \max(-s_{2j},0)\ldots \max(-s_{Kj},0)]^{T},
$$
where $j=1,2\ldots m$. Loosely speaking these two vectors compare the correaltion between all the codebook vectors and a specific {\it image descriptor} with zero or, to put in another way, compute {\it a code} for the descriptor.

Finally we form a single vector of {\it features} $\bar{f}$ by computing the maximum for each row of the matrix, whose columns are $[\hat{s}^{+}_{j}\; \hat{s}^{-}_{j}]^{T}$, resulting in the vector of the size $2K$. Roughly speaking, the last operation tries to compute the maximum linear corrrelation for a codevector and each of the image descriptors. 

The steps we described correspond in the main features to the steps used in \cite{CORNIA, Bosse}.

\subsection{Regression}

The model for an image quality metric is built using the codebook and vectors of features $\bar{f}_{k}$ computed for each image of the training data set annotated by the MOS value $l_{k}$. For that end support vector regression with either linear or non-linear kernel can be used. The result of the training is a model $l = F(\bar{f}, A)$, where by $A$ we denoted a set of implicitly presented parameters that emerged at various stages of the model construction\footnote{Note that SVR is not the only possible solution. Other approaches for building the regressor include random forests or neural networks}.

\section{Numerical experiments}
\label{sec:experiments}

In this section we present numerical analysis of the model described above as well as the results for image/video quality assessment obtained with the model. 

\subsection{Natural and synthetic data sets}

We started with the codebooks construction for which we have several options. First of all, should we use natural images or synthetic images? It seems to us the use of synthetic images has an advantage as having more control on what kind properties should be incorporated into the codebook. From the point of view of the final correlation with MOS the question of using random patches for codebooks was partially investigated in \cite{Bosse}. The authors utilized both natural images and random distributions for generating patches to be used during the codebook construction. We exploited a somewhat different approach using two types of images for building codebooks: 1) natural images; 2) synthetic images generated by means of a stochastic model.

As far as natural images is concerned, we utilized several widely known datasets which provide MOS annotations: CSIQ \cite{CSIQ}, LIVE IQA \cite{LIVEIQA} and TID2013 \cite{TID2013}, mostly consisting of natural images distorted in various ways as well as original undistorted images\footnote{Specifically, CSIQ contains 866 images and 6 types of distortions; LIVE IQA contains 779 images and 5 types of distortions; TID2013 contains 3000 images and 24 types of distortions.}. Almost all images represent some natural scenes understood broadly, i.e., contain images of people, buildings, vehicles along with nature objects. TID2013 also includes one artificial reference image which we intentionally excluded from the experiments. Each {\it distorted} image in these datasets has a MOS value that represents its subjective quality and thus can be used for training the model.

We also used datasets containing videos: LIVE VQC \cite{LIVEVQC} and LIVE VQA \cite{LIVEVQA}\footnote{LIVE VQA consists of $150$ artificially distorted videos and provides $4$ types of distortions as well as original undistorted videos. LIVE VQC consists of $585$ user generated videos that provide natural distortions.}. Unlike LIVE VQA  LIVE VQC does not contain original undistorted videos. Both datasets have MOS scores for each distorted video.

\begin{figure}
\centering
\includegraphics[width=0.93\linewidth]{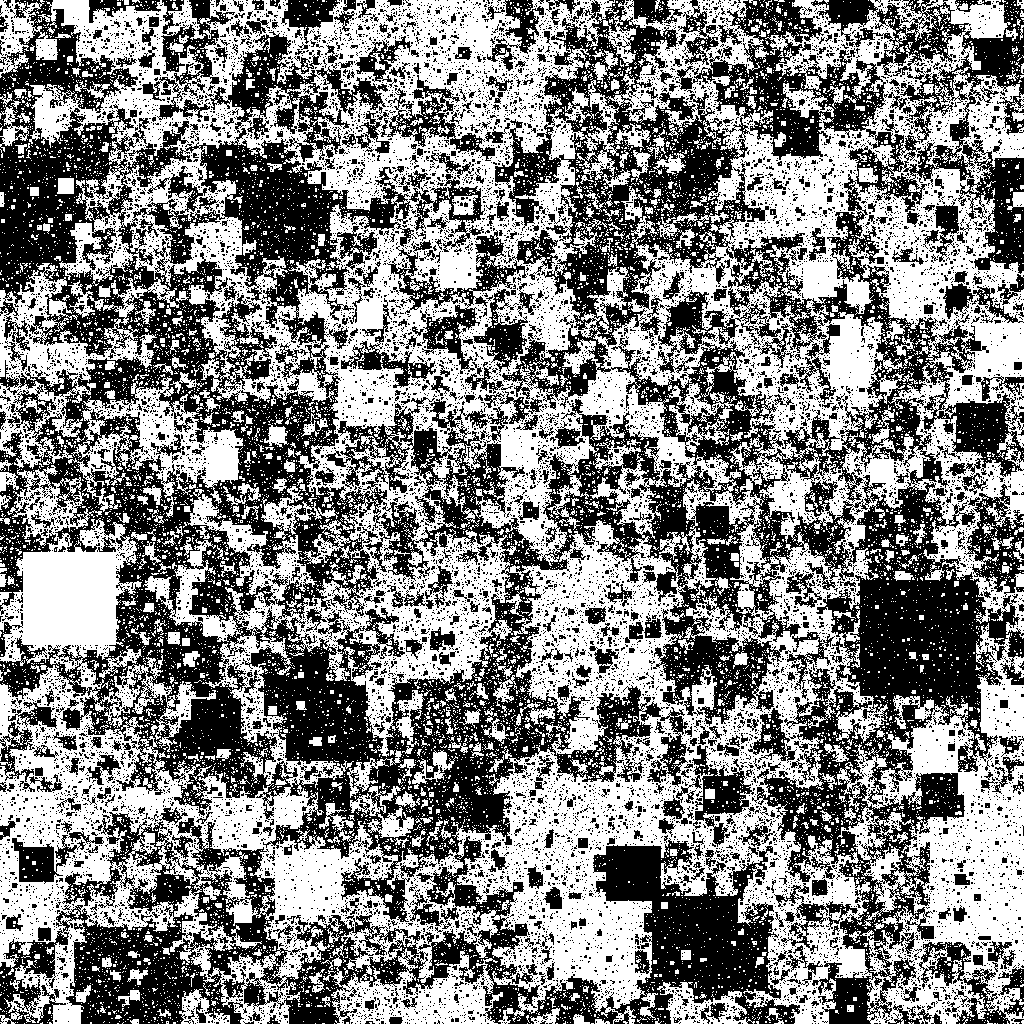}
\caption{An example of synthetic image generated by the model; for a detailed description see, e.g., \cite{Koroteev2021}. Random patches are generated whose size is randomly chosen according to some heavy tail distribution and put into the image. The process continues until the stationary state distribution is attained. In this plot a power-law distribution of patch sizes $p(x)\sim 1/x^{\gamma}$ with the exponent $\gamma=3.3$ was used as a source of patch sizes. The color of patches has two values and chooses randomly.}
\label{fig:synthetic_image}
\end{figure}

Synthetic images were generated by the model similar to that introduced back more than twenty years ago to investigate scale-invariant properties of natural images \cite{RudermanOrigin, Mumford}, see also \cite{Koroteev2021} for some additional analysis of the model. The main point of this model is its ability to reporduce scale-invariant power spectra systematically, the property observed in natural images for many years\cite{RudermanBialek}. The basic assumption for generating scale-invariant power spectra by this model is that the behaviour is determined by the edge sizes of the objects in the image. Thus the model, being sufficiently random and generic, reflects however the essential property of natural images. We applied this model to generate synthetic data sets and used them for codebook building. A typical example of an image generated by the model is given in fig. \ref{fig:synthetic_image}.

\subsection{Codebook properties}

For the computations we used image contrast $\phi(x)$, which we define in a particular point $x$ of an image as \cite{RudermanBialek, RudermanBialek2}
$$
\phi(x) = \ln\left(I / I_{0}\right),
$$
where $I(x)$ is the intensity in a given point and normalization factor $I_{0}$ is chosen for each image so as $\sum\phi(x) =0$ over the image. The intensity $I$ is measured in terms of $Y$ component of an image.

\begin{figure}
\centering
\includegraphics[width=1.01\linewidth]{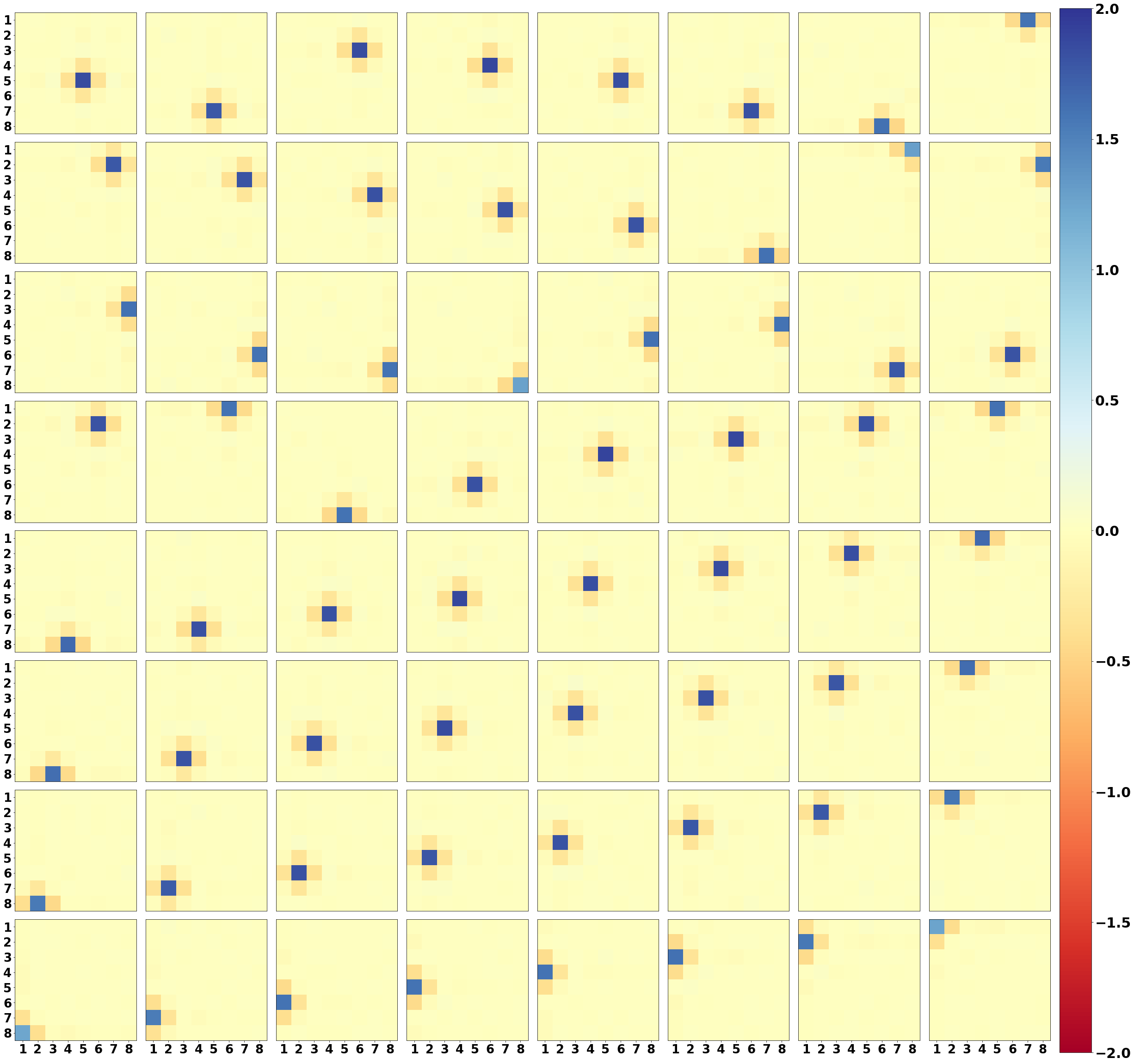}
\caption{Basis vectors of the zero component analysis (ZCA) transform (columns of matrix $U$ in (\ref{ZCA})), located in the raster order from top left to bottom right according to the decreasing eigen values (diagonal elements of $D$ in (\ref{ZCA})) for images
from CSIQ dataset. Each image was processed as explained in section \ref{sec:model}. Note the regular structure of peak locations for smaller eigen values.}
\label{fig:zca_basis}
\end{figure}

Following section \ref{sec:model} we collect a set of random patches of the size $n\times n$ pixels from the image, then we apply a whitening procedure as explained above. On fig. \ref{fig:zca_basis} we show several first columns of matrix $U$ (eigen vectors of data matrix $X$, see section \ref{sec:model}, (\ref{ZCA})) presented in the form of square patches of the size $8\times 8$ and located in the table wrt the decreasing eigen values row-wise.
The eigen vectors presented in fig. \ref{fig:zca_basis} look intuitively plausible in terms of a basis construction. For synthetic images we obtain the similar behaviour for the basis vectors (see supplemental. fig. \ref{fig:zca_basis_synthetic}) except the peak amplitudes are more uniform and of lower magnitude.

\begin{figure}
\centering
\includegraphics[width=1.02\linewidth]{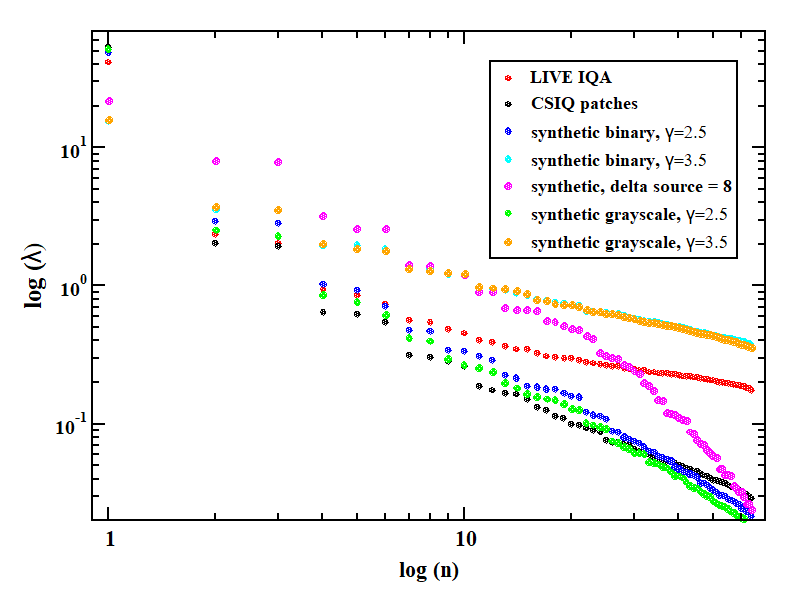}
\caption{Eigen values of the correlation matrix of a codebook in decreasing order computed for natural image data sets and synthetic data sets (see figs. \ref{fig:synthetic_image}, \ref{fig:zca_basis_synthetic}). The dimension of the space was set to $64$; note double log scale. The synthetic data include both images generated using power-law source of patches for various $\gamma$ and delta source\cite{Koroteev2021} with the fixed size patch ($=8$ in this example).}
\label{fig:zca_eigen}
\end{figure}

It also seems instructive to look at the eigen values. However, for this goal it seems more natural to use a {\it different normalization}: instead of subtracting average and division by the standard deviation for each patch, we applied the standartization for each of $64$ components of vectors. Corresponding eigenvalues of the correlation matrix are shown in fig. \ref{fig:zca_eigen}; it demonstrates a distinctive structure of eigen values both for natural and synthetic images. It is also seen that the behavior of eigen values for the synthetic images is similar for binary and greyscale cases for the same exponents of the source. This implies that the energy re-distribution over the principal components is determined primarily by the edges (gradients) of the objects presented in $8\times 8$ patches in the sample. 
The increase of the source exponent for the synthetic images (see the caption for fig. \ref{fig:synthetic_image}) results in smaller average object size and thus the eigen spectrum becomes more uniform and flattened (compare eigen spectra for $\gamma=2.5$ and $\gamma=3.5$ in fig. \ref{fig:zca_eigen}).

\begin{figure}
\centering
\includegraphics[width=1.01\linewidth]{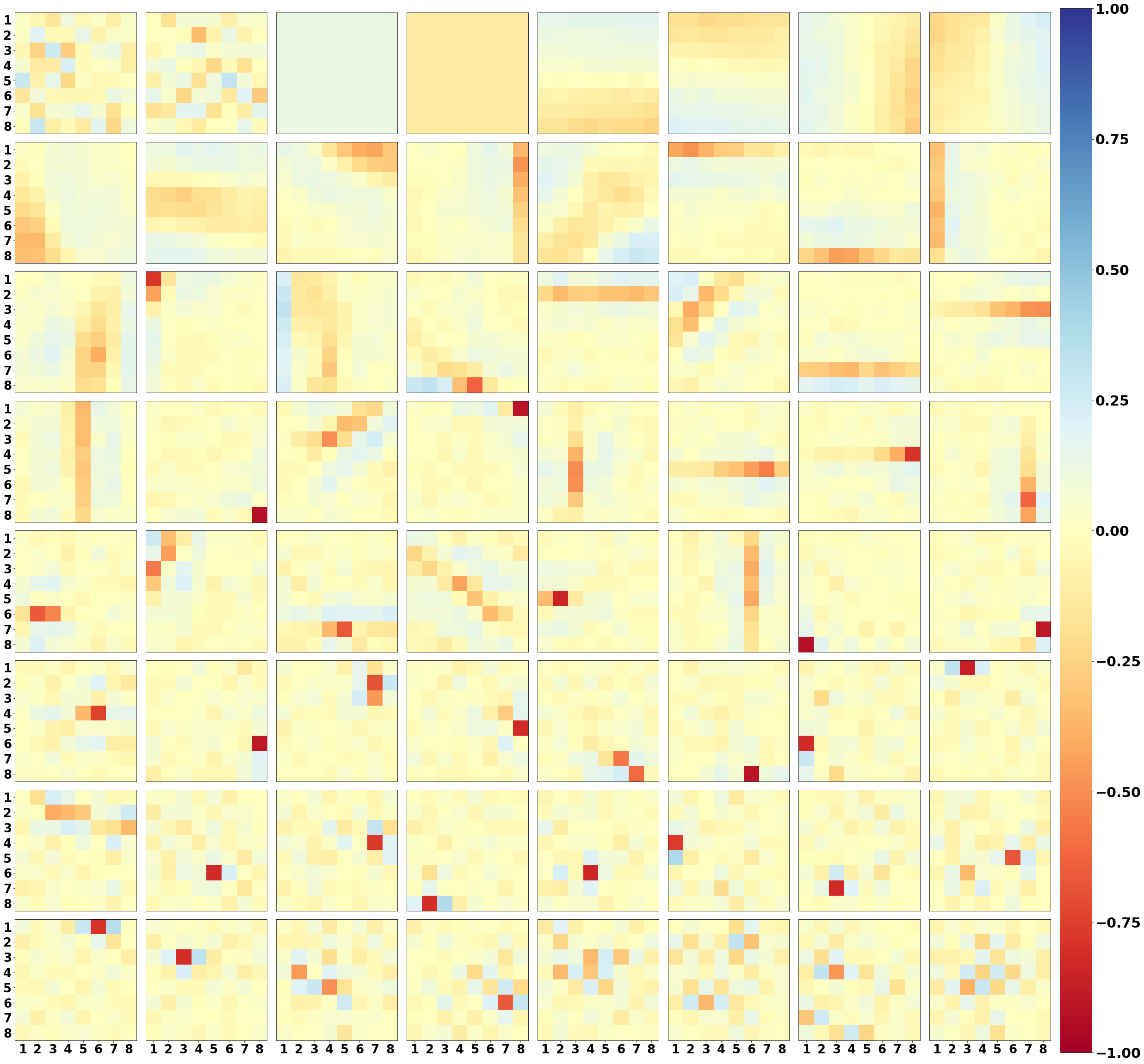}
\caption{Columns of the matrix $O$ (cf. sec. \ref{subsec:correlating}) which correspond to whitened and clustered patches of an image. The vectors (square patches of the size $8\times 8$) are presented according to the increasing average deviation of all cluster vectors from the center of the cluster additionally normalized by the length of the center of the cluster.}
\label{fig:whitened}
\end{figure}

After transforming the vectors using  basis vectors of the ZCA transform and applying K-means algorithm to obtain clusters we have the transformed vectors shown in fig. \ref{fig:whitened}.
It is seen that more dense clusters, which reflect more typical behavior of vectors, have more unformlly distributed components of the cluster centers, thus putting the center of the cluster more or less far from the axes of basis, while the increase of the variance for the vectors in the cluster results in emerging stronger peaks among the components so that these vectors tend to more correlated to the axes of the space\footnote{We also investigate the role of the whitening procedure in the codebook generation process. We implement alternative whitening method in the Fourier domain 
and generate codebooks using it instead of ZCA. Whitening is performed by performing Fourier transform on patches, computing mean of absolute value of the spectrum and dividing spectrum by this mean element-wise and taking inverse Fourier transform. We find that correlations achieved with these codebooks do not differ from correlations achieved with regular codebooks.}.

\subsection{Model set-up}

For our experiments we generated codebooks using CSIQ or LIVE IQA or the set of synthetic images. For each dataset we generate several codebooks with the number of codevectors varying from $2$ to $2048$; the patch size was chosen to be $8 \times 8$ pixels.  The number of codevectors in the codebook obviously represent the degree of details which are incorporated into the codebook.

For the most part of the experiments we used the source code written in python and provided as an opensource in \cite{Bosse} with appropriate modifications. However, as we were also interested in inference time estimates, we re-implemented the algorithm in c++ and confirmed the correlations results remain consistent between the python and c++ versions. All time measurements have been carried out for c++ implementation of the algorithm.

When building the regression model on the final stage of computations, nuSVR algorithm provided in scikit-learn library was used to predict quality scores. 

For training purposes the distorted part of each dataset was randomly split into content independent train and test subsets in proportion $\approx$ 80\% to $20\%$; $10$ random splits were carried out. For natural images the model trained on CSIQ and LIVE IQA datasets were cross-evaluated. In addition, each trained model was evaluated on TID2013 dataset. For each evaluation experiment we used the models trained on $10$ splits but the very test was fulfilled on the whole data set; the standard deviation was computed over the $10$ splits. This enabled to estimate errors and in the same time reduced the magnitude of the standard deviation that will be seen in the tables below. We also doubled the number of splits to test whether it resulted in the change in standard deviations; no significant changes were detected. Before training the features were scaled to the range $[-1, 1]$. 

For video datasets MOS values for each individual frame are not available so to train the regressors and acquire predictions we use 
the method described in \cite{VIDEVAL}:
for each video we extract first frames from the frames occurring in one second
and compute features in the usual way, then we take the average of each feature over these frames to obtain a set of features considered to be representative for the whole video. These features are then used to train SVR. Both for LIVE VQA and LIVE VQC we performed $10$ content independent splits by reference image similarly to image datasets. As for the images for LIVE VQC data set we tested the stability of the results by increasing the number of splits to $100$; changes in standard deviations remained within $0.01$.

Initially we used the default parameters $C=1$, $\eta=0.5$ provided by sklearn nuSVR function  which give the best results for the image datasets under consideration; $C=50$, $\eta=0.5$ and  $C=1000$, $\eta=0.5$ turn out to be the best for LIVE VQC and LIVE VQA correspondingly. We used RBF kernel for every dataset; this partially accounts for the improvements in terms of correlations with MOS compared to the linear kernel. However, having in mind an implementation of sufficiently fast algorithm to be used on CPU, we demonstrate that rbf kernels are suitable for that goal.

\subsection{Computational results}

Tables \ref{correlations_CSIQ_CSIQ}-\ref{correlations_CSIQ_synthetic} show mean correlations in the models trained on CSIQ with different codebooks (the codebook types are indicated in the legends for the tables). For these tables we take two values for the number of descriptors $64$ and $2048$ and vary the number of codevectors in the codebook. It is seen that overall the reduction of the number of descriptors results in more significant reduction of correlations compared to the case of the number of codevectors reduction. Also the results for TID dataset turn out to be much worse in terms of the magnitude of correlations than for other two datasets. This is accounted for by additional types of distortions presented in TID dataset, which contains roughly ten times more images and distortion types than CSIQ and LIVE IQA. 

Table \ref{correlations_CSIQ_synthetic} shows the results obtained for the codebook built using synthetic images, generated with a model described in \cite{Koroteev2021}. We note that even though the overall results are comparable for all three experiments with various codebooks, the synthetic ones demonstrate statistically significant better results on TID and, which is especially interesting, on LIVE IQA data set (for the number of descriptors equal to $2048$).

To compare what happens if the codebooks are generated using one dataset and trained using another datasaet and vice versa we can compare Table \ref{correlations_CSIQ_live} and Supplemental Table \ref{correlations_live_CSIQ}; the latter shows correlations of models trained on LIVE IQA with codebooks generated from CSIQ. Compared to Table \ref{correlations_CSIQ_live} we notice that the results vary stronger when the model is tained on LIVE IQA, which means CSIQ was able to fit better both to itself and LIVE IQA, while LIVE IQA is not that good adapted to CSIQ. Another observation, pretty obvious in terms of the training data set, is that the correlations noticeably depend on the training data set even though both CSIQ and LIVE IQA data sets are sufficiently standard to be used in IQA tasks.

\subsection{Optimizations}

In the previous sections we set forth the basic model set-up and experiments which enabled us to achieve satisfactory results in terms of the correlations with MOS. However there are other conditions for the image quality task which have to be satisfied if the algorithm is supposed to be used in real applications. Two most essential of them are inference time and the model size, which we consider now.

\begin{figure}
\centering
\includegraphics[width=1.02\linewidth]{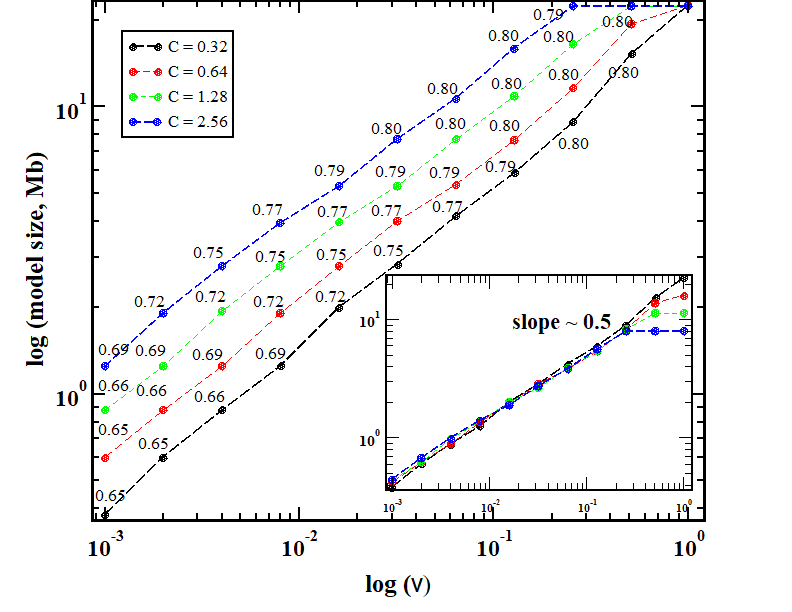} 
\caption{The dependence of the model size (in Mb) on the parameter $\nu$ ($x$-axis) and $C$ (different colors); note double log scale. In a sufficiently wide parameter regime (three order of magnitude over $\nu$) the size depends on $\nu$ according to a power law with a slope approximately $0.5$. The inset demonstrates the collapsed lines obtained by rescaling the original lines by the factor of $\sqrt{2}$. Taking into account the values for $C$ shown in the figure we conclude that the model size scales as $\sim(C\nu)^{1/2}$.}
\label{fig:model_size_vs_nu}
\end{figure}

Concerning the model size it is very desirable to have a compact model in terms of the model file size. It is clear that this size is related to the model size in RAM but for simplicity we can think of it as just the size of the file on HD. The file containing the model data is primarily occupied by the support vectors information and also contains some auxiliary information. In turn, the number of support vectors is controlled by the parameters $\nu$ and $C$ in nuSVR algorithm. This dependence is shown in  fig. \ref{fig:model_size_vs_nu}.
This plot shows that depending on the acceptable correlation level for a metric, the model size can be reduced by varying parameters $\nu$ and $C$. In practical application we thus can reduce the model size depending on the acceptable level of correlations.

Another goal we were interested to achieve was the reduction of the inference time when evaluating image quality so that the algorithm could work in real-time on CPU without a significant decrease in quality. As usual the most computationally expensive operation in the process is the multiplication of the matrix of descriptors $X$ by the matrix of codevectors $W$ (see section \ref{sec:model}). Reducing the dimensions of these matrices gives a noticeable speed-up. The dimensions correspond to the number of descriptors taken from an image and the number of codevectors in the codebook. Time depends linearly on each of these parameters, that is reducing any of them by a factor of $2$ gives us a $2x$ speed-up. We vary both of these parameters and compute correlations and timings in each case.

\subsection{Chroma information}

In the results described previously we extracted descriptors and features only from the luminance component of the image. It is possible to use the same process to extract features using chroma components. It is obvious that the chroma features can be combined with luma features in various ways. We extracted features from Y and U component of the YUV image data so that the total number of features remained the same as in the previous experiments and half of the features were obtained from the chroma image and another half from the luma image. We then trained the regressor on these joint features as above; in this experiment we did not change the codebook, i.e., the codevectors we used were constructed using only luma information. The results are shown in Table \ref{correlations_CSIQ_CSIQ_chroma}. This method achieves noticeable  improvement in terms of correlations on CSIQ dataset (Table \ref{correlations_CSIQ_CSIQ}). Note that even though the chroma information turns out to be essential, it is not sufficient for building a good model: if one constructs features using only chroma component the quality of the correlation decreases compared to only-luminance case.

\section{Experiments on video}

As it was explained above the main difference with the case of images is that for video, which can be thought of as a collection of images, the annotation with MOS is provided for all frames as a whole, rather than for individual frames. To circumvent this obstacle various methods are applied. 

\subsection{Pooling strategies}

One way to apply codebooks model to video is first, to apply feature extraction step to individual frames and then use the so called {\it average pooling} to get a set of features for the entire video. This method is used for benchmarking in \cite{VIDEVAL} for most image level algorithms including CORNIA. 

Tables \ref{tab:correlations_live_vqc} and  \ref{correlations_live_vqa} show mean correlations on LIVE VQC and LIVE VQA data sets correspondingly using average pooling; in both cases the codebooks were generated from CSIQ data set, features were computed for every frame of the video and averaged. The number of train/test splits is $100$ in both experiments.

Extracting features from every frame might be redundant since neighboring frames are usually highly correlated\footnote{We do not provide any special estimates for such correlations: they might have only theoretical interest distant from the scope of the paper. The presense of these correlations follows just from the obvious resemblance of scenes in neighbouring frames.}. We investigate the effect of sampling rate on correlation indirectly; the results are shown in Table \ref{sampling_rate}. The data shows that decreasing sampling rate does not decrease the score of the metric significantly until the rate approximately $1$ fps, after which the score starts to decline; this is an indirect evidence of correlations existing between closely located frames. The results show that the sampling rate around $1$ frame per second is reliable taking into account the measured error. Note the similar experiment was presented in \cite{TLVQM}, that demonstrated similar results for CORNIA, NIQE, BRISQUE, HIGRADE, FRIQUEE algorithms.

Another pooling strategy is {\it standard deviation pooling (std)} introduced in GMSD \cite{GMSD} and used in TLVQM \cite{TLVQM} and VIDEVAL \cite{VIDEVAL} algorithms. We calculate features for every frame as before, then for every feature we compute the mean and standard deviation across, e.g., each $1$ second segment. We concatenate the vector of means and the vector of standard deviations to get a vector of features for each segment so that the total number of features doubles compared to that without std pooling.  After that we average all segment level features to obtain the final set of features for the whole video. The difference between this method and the previous approach is not just averaging but also inclusion of the new features, standard deviations for video segments, which were then used for training SVR together with the main set of features. The resulting correlations are shown in Tables \ref{tab:std_pooling_LIVEVQC} and \ref{tab:std_pooling_KONVID}. This method allows to achieve an improvement in correlations (cf. table \ref{tab:correlations_live_vqc}). Notice that this method doubles the number of features however gives higher correlations even comparing to average pooling with the same number of features (i.e. with doubled number of codevectors).
However std pooling is slower than the average pooling since we have to compute standard deviations along with computations of features for each frame. We also tested a similar scheme but used every second and fourth frame instead of every frame to speed up the computation. We find that using every second frame still does not decrease correlations significantly, unlike using every fourth frame.

\subsection{Chroma information}

Unlike images experiments on video datasets show that using chroma component provides no improvement in terms of correlations. This can be explained by the types of distortions present in the video datasets: distortions such as contrast change that affect the chroma components are not featured in video datasets.

\subsection{Inference time reduction}

For 25 fps videos we would like to have a model that works in less than 0.04 seconds per image?. Tables \ref{timings1} and \ref{timings2} show average time for feature extraction and prediction for one image from CSIQ dataset. The experiments were performed on a machine with Intel(R) Xeon(R) Gold 6278C CPU @ 2.60GHz. The code is written in Python and can be further optimized.

\begin{table*}[htbp]
	\centering
	\caption{Average correlations on image datasets. The model trained on CSIQ dataset. Codebooks generated from CSIQ dataset; the errors (standard deviations) were estimated as follows. $10$ splits (in proportion $8:2$) of the training dataset were made and the model was trained and tested $10$ times. After that std can be estimated. The numbers in bold show the best results for $2048$ descriptors among Tables I-III.}
	\begin{tabular}{crcccccccccccc}
		\toprule
		&       & \multicolumn{4}{c}{ {CSIQ}}      & \multicolumn{4}{c}{ {LIVE IQA}}   & \multicolumn{4}{c}{ {TID2013}} \\
		\cmidrule{3-14}    \# descriptors & \# codevectors & PLCC  & STD   & SRCC  & STD   & PLCC  & STD   & SRCC  & STD   & PLCC  & STD   & SRCC  & STD \\
		\midrule
		\multirow{11}{*}{64}	& 2     & 0.40  & 0.08  & 0.26  & 0.06  & 0.40  & 0.02  & 0.36  & 0.02  & 0.31  & 0.02  & 0.18  & 0.02 \\
							    & 4     & 0.49  & 0.08  & 0.40  & 0.08  & 0.54  & 0.01  & 0.53  & 0.01  & 0.29  & 0.02  & 0.23  & 0.01 \\
							    & 8     & 0.63  & 0.06  & 0.51  & 0.07  & 0.65  & 0.01  & 0.66  & 0.01  & 0.43  & 0.02  & 0.35  & 0.01 \\
							    & 16    & 0.67  & 0.03  & 0.57  & 0.05  & 0.70  & 0.01  & 0.71  & 0.01  & 0.40  & 0.02  & 0.31  & 0.02 \\
								& 32    & 0.66  & 0.04  & 0.57  & 0.04  & 0.70  & 0.01  & 0.71  & 0.01  & 0.46  & 0.01  & 0.36  & 0.02 \\
				
		     					& 64    & 0.67  & 0.04  & 0.58  & 0.05  & 0.71  & 0.01  & 0.71  & 0.01  & 0.48  & 0.01  & 0.39  & 0.02 \\
							    & 128   & 0.68  & 0.05  & 0.60  & 0.05  & 0.72  & 0.01  & 0.72  & 0.01  & 0.47  & 0.01  & 0.36  & 0.01 \\
							    & 256   & 0.70  & 0.05  & 0.63  & 0.05  & 0.72  & 0.01  & 0.73  & 0.01  & 0.45  & 0.01  & 0.34  & 0.01 \\
							    & 512   & 0.73  & 0.05  & 0.66  & 0.05  & 0.74  & 0.01  & 0.75  & 0.01  & 0.46  & 0.01  & 0.35  & 0.01 \\
							    & 1024  & 0.72  & 0.05  & 0.65  & 0.05  & 0.74  & 0.01  & 0.75  & 0.01  & 0.48  & 0.01  & 0.37  & 0.01 \\
							    & 2048  & 0.74  & 0.04  & 0.66  & 0.05  & 0.75  & 0.01  & 0.77  & 0.01  & 0.52  & 0.01  & 0.41  & 0.01 \\
		\midrule			    
		\multirow{11}{*}{2048}  & 2     & 0.37  & 0.07  & 0.30  & 0.08  & 0.40  & 0.02  & 0.41  & 0.03  & 0.20  & 0.04  & 0.15  & 0.02 \\
		  						& 4     & 0.57  & 0.07  & 0.45  & 0.09  & 0.53  & 0.01  & 0.51  & 0.01  & 0.34  & 0.03  & 0.24  & 0.02 \\
		  						& 8     & 0.69  & 0.07  & 0.60  & 0.07  & 0.65  & 0.01  & 0.66  & 0.01  & 0.54  & 0.02  & 0.45  & 0.02 \\
		  						& 16    & 0.73  & 0.06  & 0.66  & 0.06  & 0.70  & 0.01  & 0.71  & 0.01  & 0.48  & 0.02  & 0.41  & 0.02 \\
		  						& 32    & 0.78  & 0.05  & 0.73  & 0.05  & 0.71  & 0.01  & 0.72  & 0.01  & 0.42  & 0.02  & 0.35  & 0.02 \\
	
							    & 64    & 0.76  & 0.06  & 0.71  & 0.06  & 0.72  & 0.01  & 0.73  & 0.01  & 0.55  & 0.01  & 0.48  & 0.01 \\
								& 128   & 0.76  & 0.06  & 0.72  & 0.05  & 0.71  & 0.02  & 0.73  & 0.02  & 0.53  & 0.03  & 0.46  & 0.02 \\
								& 256   & 0.78  & 0.06  & 0.73  & 0.06  & 0.73  & 0.02  & 0.74  & 0.02  & 0.50  & 0.04  & 0.42  & 0.03 \\
								& 512   & 0.77  & 0.07  & 0.73  & 0.07  & 0.73  & 0.01  & 0.75  & 0.02  & 0.50  & 0.03  & 0.43  & 0.02 \\
								& 1024  & 0.78  & 0.07  & 0.74  & 0.06  & 0.74  & 0.01  & 0.76  & 0.02  & 0.50  & 0.02  & 0.42  & 0.02 \\
								& 2048  & {\bf 0.79}  & 0.07  & {\bf 0.75}  & 0.06  & 0.76  & 0.01  & 0.79  & 0.01  & 0.52  & 0.02  & 0.44  & 0.02 \\
		\bottomrule
	\end{tabular}%
	\label{correlations_CSIQ_CSIQ}%
\end{table*}%

\begin{table*}[htbp]
	\centering
	\caption{Average correlations on image datasets. Model trained on CSIQ dataset. Codebooks generated from LIVE IQA; errors measured on $10$ splits in the same way as explained in the caption for the table \ref{correlations_CSIQ_CSIQ}. The numbers in bold show the best results for $2048$ descriptors among Tables I-III.}
	\begin{tabular}{rrrrrrrrrrrrrr}
		\toprule
		&       & \multicolumn{4}{c}{ {CSIQ}}      & \multicolumn{4}{c}{ {LIVE IQA}}   & \multicolumn{4}{c}{ {TID2013}} \\
		\cmidrule{3-14}    \# descriptors & \# codevectors & PLCC  & STD   & SRCC  & STD   & PLCC  & STD   & SRCC  & STD   & PLCC  & STD   & SRCC  & STD \\
		\midrule
		\multirow{11}{*}{64}    & 2     & 0.42  & 0.10  & 0.31  & 0.09  & 0.46  & 0.01  & 0.42  & 0.01  & 0.25  & 0.01  & 0.13  & 0.02 \\
								& 4     & 0.50  & 0.06  & 0.36  & 0.06  & 0.58  & 0.01  & 0.57  & 0.01  & 0.39  & 0.02  & 0.32  & 0.02 \\
								& 8     & 0.62  & 0.05  & 0.51  & 0.05  & 0.67  & 0.01  & 0.68  & 0.01  & 0.44  & 0.01  & 0.36  & 0.01 \\
								& 16    & 0.67  & 0.03  & 0.55  & 0.05  & 0.73  & 0.01  & 0.74  & 0.01  & 0.46  & 0.01  & 0.35  & 0.01 \\
								& 32    & 0.67  & 0.03  & 0.56  & 0.05  & 0.74  & 0.01  & 0.74  & 0.01  & 0.48  & 0.01  & 0.40  & 0.01 \\
								
								& 64    & 0.64  &  0.05  & 0.55  & 0.06  & 0.71  & 0.01  & 0.70  & 0.01  & 0.45  & 0.02  & 0.39  & 0.02 \\
								& 128   & 0.65  &  0.05  & 0.57  & 0.05  & 0.69  & 0.01  & 0.69  & 0.01  & 0.45  & 0.02  & 0.37  & 0.02 \\
								& 256   & 0.69  &  0.05  & 0.62  & 0.05  & 0.71  & 0.01  & 0.72  & 0.01  & 0.44  & 0.02  & 0.36  & 0.02 \\
								& 512   & 0.69  &  0.05  & 0.62  & 0.05  & 0.73  & 0.01  & 0.74  & 0.01  & 0.47  & 0.02  & 0.39  & 0.01 \\
								& 1024  & 0.71  & 0.05  & 0.64  & 0.06  & 0.74  & 0.01  & 0.75  & 0.01  & 0.48  & 0.01  & 0.38  & 0.01 \\
								& 2048  & 0.72  & 0.04  & 0.65  & 0.04  & 0.74  & 0.01  & 0.76  & 0.01  & 0.50  & 0.01  & 0.41  & 0.01 \\
		\midrule
		
		\multirow{11}{*}{2048}    & 2     & 0.51  & 0.11  & 0.41  & 0.10  & 0.52  & 0.02  & 0.50  & 0.03  & 0.39  & 0.01  & 0.29  & 0.01 \\
								  & 4     & 0.56  & 0.07  & 0.42  & 0.09  & 0.51  & 0.01  & 0.47  & 0.02  & 0.37  & 0.02  & 0.25  & 0.02 \\
								  & 8     & 0.64  & 0.07  & 0.55  & 0.07  & 0.66  & 0.02  & 0.66  & 0.02  & 0.42  & 0.02  & 0.33  & 0.02 \\
								  & 16    & 0.73  & 0.06  & 0.65  & 0.05  & 0.72  & 0.01  & 0.72  & 0.01  & 0.50  & 0.01  & 0.43  & 0.01 \\
								  & 32    & 0.77  & 0.05  & 0.72  & 0.05  & 0.72  & 0.01  & 0.73  & 0.01  & 0.57  & 0.01  & 0.50  & 0.02 \\
			
			 					  & 64    & 0.77  & 0.06  & 0.73  & 0.05  & 0.73  & 0.01  & 0.75  & 0.01  & 0.52  & 0.02  & 0.47  & 0.01 \\
								  & 128   & 0.76  & 0.07  & 0.72  & 0.07  & 0.69  & 0.01  & 0.72  & 0.01  & 0.57  & 0.02  & 0.51  & 0.01 \\
								  & 256   & 0.76  & 0.06  & 0.72  & 0.04  & 0.68  & 0.02  & 0.69  & 0.03  & 0.47  & 0.03  & 0.42  & 0.02 \\
								  & 512   & 0.78  & 0.07  & 0.74  & 0.07  & 0.71  & 0.01  & 0.73  & 0.01  & 0.51  & 0.02  & 0.45  & 0.01 \\
								  & 1024  & 0.79  & 0.06  & 0.75  & 0.06  & 0.71  & 0.01  & 0.73  & 0.02  & 0.50  & 0.02  & 0.43  & 0.01 \\
								  & 2048  & {\bf 0.79}  & 0.06  & {\bf 0.75}  & 0.06  & 0.74  & 0.01  & 0.76  & 0.01  & 0.50  & 0.02  & 0.43  & 0.01 \\
		\bottomrule
	\end{tabular}%
	\label{correlations_CSIQ_live}%
\end{table*}%

\begin{table*}[htbp]
	\centering
	\caption{Average correlations on image datasets.The model trained on CSIQ dataset. Codebooks generated from synthetic dataset using whitening. The std's were measured in the same as for tables \ref{correlations_CSIQ_CSIQ}, \ref{correlations_CSIQ_live}. The numbers in bold show the best results for $2048$ descriptors among Tables I-III.}
	\begin{tabular}{crcccccccccccc}
		\toprule
		&       & \multicolumn{4}{c}{ {CSIQ}}      & \multicolumn{4}{c}{ {LIVE IQA}}   & \multicolumn{4}{c}{ {TID2013}} \\
		\cmidrule{3-14}    \# descriptors & \# codevectors & PLCC  & STD   & SRCC  & STD   & PLCC  & STD   & SRCC  & STD   & PLCC  & STD   & SRCC  & STD \\
		\midrule
		
		\multirow{11}{*}{64}    & 2     & 0.26  & 0.07  & 0.23  & 0.06  & 0.32  & 0.02  & 0.32  & 0.03  & 0.12  & 0.02  & 0.07  & 0.01 \\
							    & 4     & 0.48  & 0.08  & 0.32  & 0.08  & 0.43  & 0.01  & 0.36  & 0.02  & 0.20  & 0.01  & 0.11  & 0.01 \\
							    & 8     & 0.56  & 0.05  & 0.44  & 0.05  & 0.54  & 0.01  & 0.51  & 0.01  & 0.35  & 0.02  & 0.27  & 0.02 \\
							    & 16    & 0.63  & 0.05  & 0.47  & 0.06  & 0.67  & 0.01  & 0.67  & 0.01  & 0.42  & 0.01  & 0.29  & 0.02 \\
							    & 32    & 0.67  & 0.06  & 0.55  & 0.07  & 0.71  & 0.01  & 0.71  & 0.01  & 0.40  & 0.01  & 0.27  & 0.02 \\
		
		     					& 64    & 0.67  & 0.05  & 0.57  & 0.06  & 0.71  & 0.01  & 0.71  & 0.01  & 0.42  & 0.02  & 0.29  & 0.02 \\
								& 128   & 0.65  & 0.06  & 0.54  & 0.07  & 0.71  & 0.01  & 0.71  & 0.01  & 0.42  & 0.01  & 0.31  & 0.01 \\
								& 256   & 0.69  & 0.05  & 0.59  & 0.06  & 0.71  & 0.01  & 0.71  & 0.01  & 0.41  & 0.02  & 0.29  & 0.02 \\
								& 512   & 0.68  & 0.06  & 0.59  & 0.06  & 0.72  & 0.01  & 0.72  & 0.01  & 0.44  & 0.01  & 0.31  & 0.01 \\
								& 1024  & 0.69  & 0.06  & 0.59  & 0.07  & 0.73  & 0.01  & 0.74  & 0.01  & 0.45  & 0.01  & 0.32  & 0.01 \\
								& 2048  & 0.71  & 0.05  & 0.62  & 0.06  & 0.74  & 0.01  & 0.76  & 0.01  & 0.48  & 0.01  & 0.35  & 0.02 \\
		\midrule
		\multirow{11}{*}{2048}  & 2     & 0.33  & 0.12  & 0.20  & 0.10  & 0.37  & 0.02  & 0.34  & 0.02  & 0.19  & 0.01  & 0.15  & 0.02 \\
								& 4     & 0.65  & 0.03  & 0.52  & 0.06  & 0.59  & 0.01  & 0.55  & 0.01  & 0.33  & 0.02  & 0.24  & 0.04 \\
								& 8     & 0.70  & 0.06  & 0.62  & 0.07  & 0.62  & 0.02  & 0.61  & 0.02  & 0.49  & 0.02  & 0.41  & 0.02 \\
							    & 16    & 0.67  & 0.05  & 0.57  & 0.06  & 0.70  & 0.01  & 0.70  & 0.02  & 0.42  & 0.03  & 0.38  & 0.03 \\
								& 32    & 0.73  & 0.05  & 0.64  & 0.05  & 0.74  & 0.01  & 0.75  & 0.01  & 0.52  & 0.02  & 0.46  & 0.02 \\
		
								& 64    & 0.74  & 0.05  & 0.66  & 0.06  & 0.74  & 0.01  & 0.74  & 0.01  & 0.56  & 0.01  & 0.51  & 0.01 \\
								& 128   & 0.71  & 0.08  & 0.65  & 0.07  & 0.74  & 0.01  & 0.75  & 0.01  & 0.64  & 0.02  & 0.60  & 0.02 \\
								& 256   & 0.76  & 0.07  & 0.70  & 0.07  & 0.76  & 0.01  & 0.77  & 0.01  & 0.62  & 0.01  & 0.57  & 0.01 \\
								& 512   & 0.74  & 0.08  & 0.69  & 0.06  & 0.78  & 0.01  & 0.79  & 0.01  & 0.65  & 0.01  & 0.58  & 0.01 \\
								& 1024  & 0.74  & 0.09  & 0.68  & 0.08  & 0.79  & 0.01  & 0.80  & 0.01  & 0.65  & 0.01  & 0.60  & 0.01 \\
								& 2048  & 0.76  & 0.07  & 0.70  & 0.06  & {\bf 0.80}  & 0.01  & {\bf 0.82}  & 0.01  & {\bf 0.66}  & 0.01  & {\bf 0.60}  & 0.01 \\
		\bottomrule
	\end{tabular}%
	\label{correlations_CSIQ_synthetic}%
\end{table*}%

\begin{table*}[htbp]
	\centering
	\caption{Average correlations with MOS on image datasets. The model was trained on CSIQ dataset; codebooks generated from CSIQ dataset.  Features for this experiment were generated using information both from luma ($1/2$) and chroma ($1/2$) images so that the resulting number of features was the same as for the experiments which used only luma information.}
	\begin{tabular}{crcccccccccccc}
		\toprule
		&       & \multicolumn{4}{c}{ {CSIQ}}      & \multicolumn{4}{c}{ {LIVE IQA}}   & \multicolumn{4}{c}{ {TID2013}} \\
		\cmidrule{3-14}    \# descriptors & \# codevectors & PLCC  & STD   & SRCC  & STD   & PLCC  & STD   & SRCC  & STD   & PLCC  & STD   & SRCC  & STD \\
		\midrule
		\multirow{11}{*}{64}    & 2     & 0.54  & 0.09  & 0.41  & 0.08  & 0.53  & 0.02  & 0.51  & 0.02  & 0.38  & 0.02  & 0.25  & 0.01 \\
							    & 4     & 0.58  & 0.06  & 0.48  & 0.06  & 0.64  & 0.01  & 0.63  & 0.01  & 0.41  & 0.01  & 0.32  & 0.01 \\
							    & 8     & 0.75  & 0.05  & 0.70  & 0.07  & 0.73  & 0.01  & 0.74  & 0.01  & 0.50  & 0.01  & 0.43  & 0.01 \\
							    & 16    & 0.74  & 0.06  & 0.68  & 0.07  & 0.78  & 0.01  & 0.79  & 0.01  & 0.54  & 0.01  & 0.46  & 0.01 \\
							    & 32    & 0.76  & 0.06  & 0.71  & 0.07  & 0.76  & 0.01  & 0.77  & 0.01  & 0.53  & 0.01  & 0.46  & 0.01 \\
							    & 64    & 0.78  & 0.04  & 0.72  & 0.05  & 0.80  & 0.01  & 0.81  & 0.01  & 0.55  & 0.01  & 0.49  & 0.01 \\
							    & 128   & 0.79  & 0.04  & 0.75  & 0.05  & 0.80  & 0.01  & 0.81  & 0.01  & 0.54  & 0.02  & 0.47  & 0.01 \\
							    & 256   & 0.80  & 0.05  & 0.76  & 0.06  & 0.81  & 0.01  & 0.82  & 0.01  & 0.54  & 0.01  & 0.47  & 0.02 \\
							    & 512   & 0.82  & 0.05  & 0.78  & 0.06  & 0.83  & 0.01  & 0.84  & 0.01  & 0.55  & 0.01  & 0.48  & 0.01 \\
							    & 1024  & 0.83  & 0.04  & 0.79  & 0.05  & 0.83  & 0.01  & 0.84  & 0.01  & 0.55  & 0.01  & 0.48  & 0.01 \\
							    & 2048  & 0.83  & 0.04  & 0.80  & 0.05  & 0.83  & 0.01  & 0.85  & 0.01  & 0.57  & 0.01  & 0.50  & 0.01 \\
		\midrule
		\multirow{11}{*}{2048}    & 2     & 0.51  & 0.14  & 0.43  & 0.10  & 0.59  & 0.02  & 0.57  & 0.02  & 0.35  & 0.03  & 0.23  & 0.02 \\
								  & 4     & 0.69  & 0.09  & 0.61  & 0.09  & 0.70  & 0.01  & 0.69  & 0.01  & 0.50  & 0.03  & 0.43  & 0.03 \\
								  & 8     & 0.79  & 0.05  & 0.73  & 0.06  & 0.76  & 0.01  & 0.77  & 0.01  & 0.58  & 0.02  & 0.52  & 0.02 \\
								  & 16    & 0.81  & 0.04  & 0.77  & 0.04  & 0.75  & 0.01  & 0.76  & 0.01  & 0.44  & 0.02  & 0.37  & 0.02 \\
								  & 32    & 0.83  & 0.04  & 0.79  & 0.04  & 0.77  & 0.01  & 0.78  & 0.01  & 0.42  & 0.03  & 0.36  & 0.02 \\
								  & 64    & 0.81  & 0.07  & 0.78  & 0.06  & 0.78  & 0.01  & 0.79  & 0.01  & 0.52  & 0.02  & 0.44  & 0.02 \\
								  & 128   & 0.83  & 0.06  & 0.80  & 0.06  & 0.82  & 0.01  & 0.83  & 0.01  & 0.48  & 0.02  & 0.42  & 0.02 \\
								  & 256   & 0.82  & 0.06  & 0.80  & 0.06  & 0.82  & 0.01  & 0.83  & 0.01  & 0.47  & 0.03  & 0.40  & 0.02 \\
								  & 512   & 0.84  & 0.06  & 0.81  & 0.06  & 0.83  & 0.01  & 0.84  & 0.01  & 0.49  & 0.01  & 0.42  & 0.02 \\
								  & 1024  & 0.84  & 0.06  & 0.81  & 0.06  & 0.84  & 0.00  & 0.86  & 0.00  & 0.51  & 0.01  & 0.44  & 0.01 \\
								  & 2048  & 0.85  & 0.06  & 0.82  & 0.06  & 0.85  & 0.00  & 0.86  & 0.00  & 0.53  & 0.01  & 0.45  & 0.01 \\
		
		\bottomrule
	\end{tabular}%
	\label{correlations_CSIQ_CSIQ_chroma}%
\end{table*}%

\begin{table}[htbp]
	\centering
	\caption{Average correlations on LIVE VQC with the application on average pooling. The model trained on LIVE VQC dataset. Codebooks generated from CSIQ dataset.}
	\begin{tabular}{crcccc}
		\toprule
		&       & \multicolumn{4}{c}{ {LIVE VQC}} \\
		\cmidrule{3-6}    \multicolumn{1}{c}{\# descriptors} & \multicolumn{1}{c}{\# codevectors} & \multicolumn{1}{c}{PLCC} & \multicolumn{1}{c}{STD} & \multicolumn{1}{c}{SRCC} & \multicolumn{1}{c}{STD} \\
		\midrule
		
		\multirow{11}{*}{64}     & 2     & 0.22  & 0.08  & 0.23  & 0.08 \\
							    & 4     & 0.47  & 0.06  & 0.43  & 0.05 \\
							    & 8     & 0.58  & 0.04  & 0.56  & 0.04 \\
							    & 16    & 0.59  & 0.04  & 0.57  & 0.04 \\
							    & 32    & 0.62  & 0.05  & 0.58  & 0.06 \\
							    & 64    & 0.62  & 0.05  & 0.59  & 0.06 \\
							    & 128   & 0.63  & 0.05  & 0.60  & 0.05 \\
							    & 256   & 0.64  & 0.04  & 0.60  & 0.05 \\
							    & 512   & 0.64  & 0.04  & 0.61  & 0.04 \\
							    & 1024  & 0.64  & 0.04  & 0.61  & 0.04 \\
							    & 2048  & 0.64  & 0.04  & 0.61  & 0.04 \\
		\midrule
		
		\multirow{11}{*}{2048} & 2     & 0.26  & 0.07  & 0.26  & 0.06 \\
							  & 4     & 0.61  & 0.03  & 0.56  & 0.04 \\
							  & 8     & 0.65  & 0.05  & 0.59  & 0.05 \\
							  & 16    & 0.68  & 0.05  & 0.65  & 0.04 \\
							  & 32    & 0.70  & 0.04  & 0.66  & 0.06 \\
		 					  & 64    & 0.72  & 0.05  & 0.69  & 0.05 \\
							  & 128   & 0.73  & 0.04  & 0.70  & 0.04 \\
					  		  & 256   & 0.72  & 0.03  & 0.68  & 0.04 \\
					  		  & 512   & 0.73  & 0.03  & 0.68  & 0.03 \\
					  		  & 1024  & 0.73  & 0.03  & 0.68  & 0.04 \\
					  		  & 2048  & 0.73  & 0.04  & 0.69  & 0.04 \\
		\bottomrule
	\end{tabular}%
	\label{tab:correlations_live_vqc}%
\end{table}%

\begin{table}[htbp]
	\centering
	\caption{Mean correlations on LIVE VQA with application of average pooling. Model trained on LIVE VQA dataset. Codebooks generated from CSIQ dataset.}
	\begin{tabular}{crcccc}
		\toprule
		&       & \multicolumn{4}{c}{ {LIVE VQA}} \\
		\cmidrule{3-6}    \# descriptors & \# codevectors & PLCC  & STD   & SRCC  & STD \\
		\midrule
		\multirow{11}{*}{64}    & 2     & -0.1  & 0.2   & -0.1  & 0.2 \\
							    & 4     & 0.2   & 0.2   & 0.2   & 0.2 \\
							    & 8     & 0.0   & 0.1   & 0.0   & 0.1 \\
							    & 16    & 0.0   & 0.2   & 0.0   & 0.2 \\
							    & 32    & 0.2   & 0.2   & 0.2   & 0.2 \\
							    & 64    & 0.2   & 0.1   & 0.1   & 0.1 \\
							    & 128   & 0.2   & 0.1   & 0.2   & 0.1 \\
							    & 256   & 0.4   & 0.1   & 0.3   & 0.1 \\
							    & 512   & 0.3   & 0.1   & 0.3   & 0.1 \\
							    & 1024  & 0.3   & 0.1   & 0.3   & 0.1 \\
							    & 2048  & 0.3   & 0.1   & 0.3   & 0.1 \\
		\midrule
		\multirow{11}{*}{2048}    & 2     & -0.1  & 0.1   & 0.0   & 0.1 \\
								  & 4     & 0.0   & 0.1   & 0.0   & 0.1 \\
								  & 8     & -0.1  & 0.2   & -0.1  & 0.2 \\
								  & 16    & 0.2   & 0.2   & 0.1   & 0.2 \\
								  & 32    & 0.4   & 0.1   & 0.4   & 0.2 \\
								  & 64    & 0.3   & 0.2   & 0.2   & 0.2 \\
								  & 128   & 0.3   & 0.2   & 0.3   & 0.2 \\
								  & 256   & 0.4   & 0.2   & 0.4   & 0.2 \\
								  & 512   & 0.5   & 0.1   & 0.4   & 0.2 \\
								  & 1024  & 0.5   & 0.2   & 0.5   & 0.2 \\
								  & 2048  & 0.5   & 0.2   & 0.5   & 0.2 \\
		\bottomrule
	\end{tabular}%
	\label{correlations_live_vqa}%
\end{table}%

\begin{table}[htbp]
	\centering
	\caption{Timings for feature extraction. Mean inference time for one image in milliseconds evaluated on images from CSIQ. Feature extraction includes two main computations: 1) feature encoding, i.e., descriptors and codevectors matrix multiplication and $\max$ operation; 2) additional computations including descriptors extraction and sample standardization.}
	\begin{tabular}{crllcc}
		\toprule
		\multicolumn{1}{c}{\# descriptors} & \multicolumn{1}{c}{\# codevectors} & \multicolumn{1}{c}{Features} & \multicolumn{1}{c}{STD}  & \multicolumn{1}{c}{Additional} & \multicolumn{1}{c}{STD} \\
		& &enc. time & &cmpns (ms)\\
		\midrule
		\multirow{11}{*}{64} & 2     & 0.08 & 0.01 & 1.1 & 0.1 \\
							& 4     & 0.08 & 0.01 & 1.1 & 0.1 \\
							& 8     & 0.08 & 0.01 & 1.2 & 0.1 \\
							& 16    & 0.09 & 0.01 & 1.3 & 0.2 \\
							& 32    & 0.15 & 0.01 & 1.1 & 0.1 \\
							& 64    & 0.17 & 0.02 & 1.0 & 0.1 \\
							& 128   & 0.22 & 0.03 & 1.1 & 0.1 \\
							& 256   & 0.4 & 0.1 & 1.1 & 0.1 \\
							& 512   & 0.8 & 0.1 &1.2 & 0.1 \\
							& 1024  & 1.7 & 0.1 & 1.0 & 0.1 \\
							& 2048  & 2.7 & 0.1 & 1.0 & 0.1 \\

		\midrule
		
		\multirow{11}{*}{2048}  &   2     & 0.4 & 0.1 & 1.6 & 0.1 \\
								&	4     & 0.6 & 0.1 & 1.7 & 0.3 \\
								&	8     & 0.6 & 0.3 & 1.7 & 0.2 \\
								&	16    & 0.7 & 0.1 & 1.7 & 0.2 \\
								&	32    & 1.4 & 0.1 & 1.7 & 0.1 \\
								&	64    & 2.6 & 0.1 & 1.1 & 0.1 \\
								&	128   & 5.3 & 0.2 & 1.2 & 0.1 \\
								&	256   & 9.4 & 0.4 & 1.3 & 0.1 \\
								&	512   & 18.5 & 0.4 & 1.4 & 0.1 \\
								&	1024  & 37.2 & 0.3 & 1.5 & 0.1 \\
								&	2048  & 68.0 & 0.8 & 2.0 & 0.1 \\

		\bottomrule
	\end{tabular}%
	
	\label{timings1}%
\end{table}%

\begin{table}[htbp]
	\centering
	\caption{Timings for prediction. Mean time in {\it milliseconds} for one image evaluated on images from CSIQ. The time for prediction does not depend on the number of descriptors, but only on the number of codevectors,}
	\begin{tabular}{rll}
		\toprule
		\multicolumn{1}{c}{\# codevectors} & \multicolumn{1}{c}{Prediction}  & \multicolumn{1}{c}{STD} \\
		\midrule
		2     & 0.0091 & 0.0004 \\
		4     & 0.0118 & 0.0008 \\
		8     & 0.0158 & 0.0009 \\
		16    & 0.026  & 0.001 \\
		32    & 0.047  & 0.002 \\
		64    & 0.093  & 0.003 \\
		128   & 0.199 & 0.008 \\
		256   & 0.43   & 0.02 \\
		512   & 0.89   & 0.03 \\
		1024  & 1.83   & 0.06 \\
		2048  & 3.7    & 0.1 \\
		\bottomrule
	\end{tabular}%
	\label{timings2}%
	
\end{table}%

\begin{table*}[htbp]
	\centering
	\caption{Correlations with MOS on LIVE VQC depending on the frame sampling rate for pooling. \# codevectors and descriptors is 2048, the codebook generated from CSIQ.}
	\begin{tabular}{lcccccc}
		\toprule
		 \multicolumn{1}{l}{{sampling rate (frame / sec.)}} & \multicolumn{1}{c}{{plcc mean}} & \multicolumn{1}{c}{{plcc std}} & \multicolumn{1}{c}{{plcc median}} & \multicolumn{1}{c}{{srcc mean}} & \multicolumn{1}{c}{{srcc std}} & \multicolumn{1}{c}{{srcc median}} \\
		 \midrule
		 30 & 0.73  & 0.04  & 0.73  & 0.69  & 0.05  & 0.69 \\
		 16   & 0.73  & 0.04  & 0.74  & 0.70  & 0.05  & 0.70 \\
		 8   & 0.73  & 0.04  & 0.73  & 0.70  & 0.05  & 0.70 \\
		 4     & 0.72  & 0.04  & 0.73  & 0.68  & 0.05  & 0.69 \\
		 2     & 0.73  & 0.04  & 0.73  & 0.69  & 0.04  & 0.69 \\
		 1     & 0.73  & 0.03  & 0.74  & 0.70  & 0.04  & 0.70 \\
		 0.5  & 0.71  & 0.04  & 0.72  & 0.67  & 0.05  & 0.68 \\
		 0.25 & 0.70  & 0.04  & 0.70  & 0.66  & 0.05  & 0.66 \\
		 1 frame per video & 0.65  & 0.05  & 0.65  & 0.62  & 0.05  & 0.62 \\
		 \bottomrule
	\end{tabular}%
	\label{sampling_rate}%
\end{table*}%

\begin{table*}[htbp]
	\centering
	\caption{Correlations with MOS on LIVE VQC data set using standard deviation pooling (the codebook generated from CSIQ). Note that the number of features is 2 times higher than that of average pooling}
	\begin{tabular}{rrlrrrrrr}
		\toprule
		\multicolumn{1}{c}{{ \# descriptors}} & \multicolumn{1}{c}{{ \# codevectors}} & \multicolumn{1}{l}{{\# of frames used}} & \multicolumn{1}{c}{{plcc mean}} & \multicolumn{1}{c}{{plcc std}} & \multicolumn{1}{c}{{plcc median}} & \multicolumn{1}{c}{{srcc mean}} & \multicolumn{1}{c}{{srcc std}} & \multicolumn{1}{c}{{srcc median}} \\
		\midrule
		 &     & every frame & 0.68  & 0.05  & 0.69  & 0.65  & 0.05  & 0.65  \\
		64    & 64    & every second frame & 0.68  & 0.05  & 0.69  & 0.65  & 0.05  & 0.65  \\
		&     & every fourth frame & 0.66  & 0.05  & 0.67  & 0.62  & 0.05  & 0.62  \\
		\midrule
		  &   & every frame & 0.76  & 0.03  & 0.77  & 0.73  & 0.04  & 0.74  \\
		2048  & 2048  & every second frame & 0.76  & 0.03  & 0.76  & 0.73  & 0.04  & 0.74  \\
		  &   & every fourth frame & 0.75  & 0.03  & 0.76  & 0.73  & 0.04  & 0.73  \\
		\midrule
		  &   & every frame & 0.77  & 0.03  & 0.77  & 0.74  & 0.04  & 0.75  \\
		4096  & 4096  & every second frame & 0.77  & 0.03  & 0.77  & 0.74  & 0.04  & 0.75  \\
		 &   & every fourth frame & 0.76  & 0.03  & 0.77  & 0.74  & 0.04  & 0.75  \\
		\bottomrule
	\end{tabular}%
	\label{tab:std_pooling_LIVEVQC}%
\end{table*}%

\begin{table*}[htbp]
	\centering
	\caption{Correlations with MOS on KONVID data set using standard deviation pooling (the codebook generated from CSIQ). Note that the number of features is 2 times higher than that of average pooling}
	\begin{tabular}{rrlrrrrrr}
		 \toprule
		 \multicolumn{1}{c}{{ \# descriptors}} & \multicolumn{1}{c}{{ \# codevectors}} & \multicolumn{1}{c}{{ \# of frames used}} & \multicolumn{1}{c}{{plcc mean}} & \multicolumn{1}{c}{{plcc std}} & \multicolumn{1}{c}{{plcc median}} & \multicolumn{1}{c}{{srcc mean}} & \multicolumn{1}{c}{{srcc std}} & \multicolumn{1}{c}{{srcc median}} \\
		 \midrule
		  &   & every frame & 0.64  & 0.04  & 0.64  & 0.64  & 0.03  & 0.64  \\
		 64    & 64    & every second frame & 0.63  & 0.03  & 0.63  & 0.63  & 0.03  & 0.63  \\
		 &    & every fourth frame & 0.62  & 0.03  & 0.63  & 0.63  & 0.03  & 0.62  \\
		 \midrule
		 &   & every frame & 0.77  & 0.02  & 0.77  & 0.77  & 0.03  & 0.77  \\
		 2048  & 2048  & every second frame & 0.77  & 0.02  & 0.77  & 0.77  & 0.03  & 0.77  \\
		 &  & every fourth frame & 0.76  & 0.02  & 0.76  & 0.77  & 0.03  & 0.77  \\
		 \midrule
		 &  & every frame & 0.78  & 0.02  & 0.78  & 0.78  & 0.02  & 0.78  \\
		 4096  & 4096  & every second frame & 0.78  & 0.02  & 0.78  & 0.78  & 0.02  & 0.78  \\
		 &  & every fourth frame & 0.78  & 0.02  & 0.77  & 0.78  & 0.02  & 0.78  \\
		 \bottomrule
	\end{tabular}%
	\label{tab:std_pooling_KONVID}%
\end{table*}%

We investigate the performance of the algorithm depending on the resolution of the video. We use 3 subsets of LIVE VQC consisting of videos of $1920\times1080$, $1280\times720$ and $404\times720$ pixels. The subsets contain 110, 316 and 119 videos respectively. We split each subset into 80\%/20\% train/test sets and train and test SVR on them. The results are shown in Table \ref{tab:LIVEVQC_resolution}.
Surprisingly we find that the correlations are similar for subsets of different resolutions as one could expect assuming the high resolution images contain more details (more edges). This suggests that correlation does not depend on the resolution only on the number of descriptors. For many IQA and VQA algorithms speed is highly dependent on resolution and often scales poorly when resolution increases, while our algorithm does not change in any way for videos of different resolutions and does not use any information about resolution. 

\subsection{Dependence on bitrate}

\begin{figure}
\centering
\includegraphics[width=1.03\linewidth]{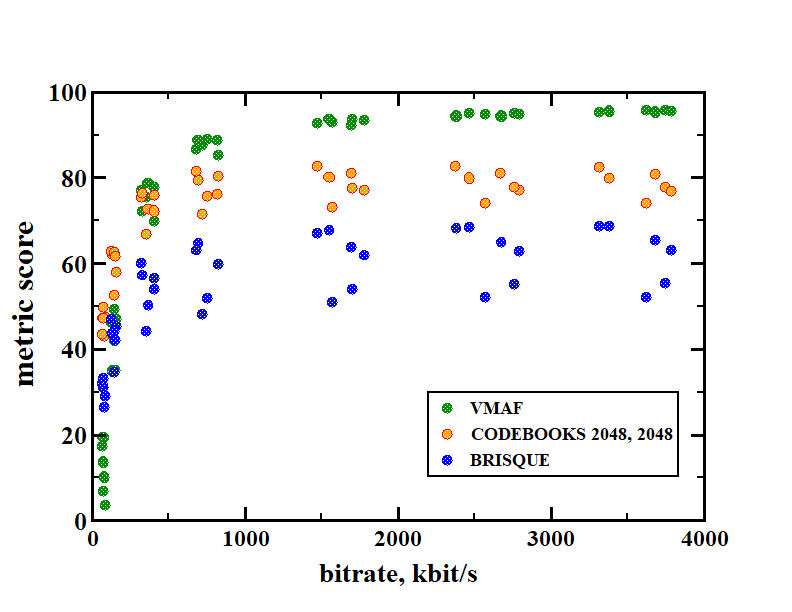}
\caption{The codebooks model trained on LIVE VQC data set to test on video data. There are clearly seen six groups of points. This is related to the structure of the test set: it contains multiple videos unified into groups with similar bitrates. Both non-reference metrics (ours and BRISQUE) show noticeably smaller span over the magnitude of the metric score compared to VMAF. For the codebooks algorithm the metric is restricted roughly by $80$ from the top and by $40$ from the bottom. The parameters for the codebooks alorithms are (codevectors/descriptors)$=2048/2048$}
\label{fig:bitrate}
\end{figure}

We tested our models on a dataset consisting of video conferencing videos that feature distortions caused by various levels of video compression. While testing on this dataset it was noticed that our metric score for every video demonstrated relatively narrow magnitude range (approximately from $20$ to $60$), while the desired output of the model should cover the whole range from $0$ to $100$. When we say the range is relatively narrow, we should realize that this relative comparison if quite inadequate: the metrics have different normalizations so it may be they are just incomparable even though they all can be dimensionless. However, we still would be interested in trying to look how the metric range wrt. bitrate varies depending on other parameters. The narrow range of the metric score would imply a weak sensitivity of the metric to the distortions and can be caused by various reasons, e.g., the very data set may cover a small region in the feature space. To see the narrow range of the score more clearly we compared our metric score with another non-reference metric, BRISQUE\cite{brisque}, which demonstrated good results in the tests and also with VMAF\cite{VMAF}, the full-reference metric; the results are shown in fig. \ref{fig:bitrate}. 
\begin{figure}
\centering
\includegraphics[width=1.03\linewidth]{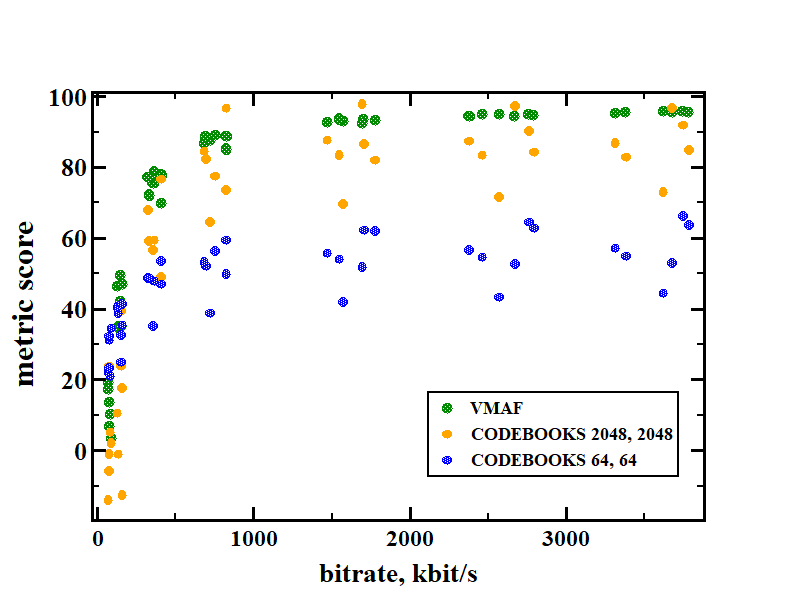}
\caption{The codebooks model trained on VQEGHD3 data set to test on video data. Some effects are mentioned in the caption to \ref{fig:bitrate} and are not repeated here. For the codebooks algorithm with the parameters  (codevectors/descriptors)$=2048/2048$ the metric spans over the range $>100$, which is comparable to VMAF. For comparison the model with much smalle numbers  (codevectors/descriptors)$=64/64$ is provided. Note, the metric score can take on negative values; this issues can be corrected by rescaling (not provided here).}
\label{fig:bitrate2}
\end{figure}
It is seen that both non-reference metrics demonstrate similar behaviour in terms of the metric magnitude and sensitivity to the bitrate though even for these two metrics the normalizations can be different. We may suggest the non-reference metrics behave similarly compared to the full-reference metric. To investigate to what extent this behaviour depends on the training set we re-trained the model, using a more extensive data set VQEGHD3 \cite{VQEGHD3}; the results are shown in fig. \ref{fig:bitrate2}. It is seen that for the same parameters of the codebook model the range of the metric score turns out to be noticeably wider than in the previous experiment and close to VMAF. The results suggest there is a dependence of the metric sensitivity to bitrate on the data set. It is noticeable that the non-reference metric behaves more noisy compared to VMAF for each group of points. In addition, we tried to analyze how the sensitivity of the metric depends on the number of codevectors and descriptors. In fig. \ref{fig:bitrate2} we provide comparison for codebooks with two parameters: the number of codevectors and the number of descriptors. The results confirm the intuitive conclusion: the system becomes more sensitive when increasing the number of parameters (features), so that for the number of codevectors/descriptors$=2048/2048$ the metric attains almost the same values as VMAF while for the number of codevectors/descriptors$=64/64$ the metric remains insensitive even for the large training data set.

In terms of the practical correction of the issue we discussed above we can try to rescale the metric by including information about bitrate into the model by multiplying the output of SVR by, e.g., $\frac{c \sqrt b}{(k+\sqrt b)}$ where parameters $c$ and $k$ are selected so the interval of outputs is from $0$ to $100$ and $b$ is the bitrate; $k$ is obviously chosen as a square root of some bitrate value. This method expands the range of outputs but the resulting metric might not reflect the true MOS and uses the bitrate information that might not be available in some situations.


\subsection{Performance comparisons}

In previous sections we analyzed the results obtained for the CORNIA-like codebooks algorithm for various types of codebooks, and training/test data sets. Here we provide some comparisons to the results collected from the literature. Tables \ref{comparison} and \ref{comparison_video} summarize the performance of our method in comparison to other IQA and VQA algorithms. Even though in terms of absolute values of correlations our method does not always provide the highest value but it usually remains within the statistical significance of the measurement. These results are observed both for images and video data. As we noticed earlier, the addition of chroma information evidently improves the correlations in the case of image quality assessment; for video the effect is too small if any. For video data std pooling method enables to achieve some improvement in correlations.

We also provide results on how synthetic codebooks affect the correlations for various data sets. These results are presents in Table \ref{tab:synthetic_codebooks}. It is seen that for some experiments the synthetic codebooks demonstrate superior correlations over other methods, even though the difference can be of the order of standard deviation. This, however, may depend on how to measure the latter. Our method of measurement is quite rough and explained in some details in the caption of Table \ref{tab:synthetic_codebooks} as well as in the captions to some other tables. It seems interesting to note that if uncorrelated noise is used for codebook building the results do not demontrate a big difference whith natural images of synthetic ones; similar observations were earlier made in \cite{Bosse}. It seems quite natural that for the most part of experiments the procedure of whitening does not affect the codebooks constructed from noise. In the same time, for the synthetic codebooks the removal of ZCA may provide better results in terms of correaltions. A more detailed analysis is left as a topic for further research.

\begin{table*}[htbp]
	\centering
	\caption{Correlations with MOS on LIVE VQC depending on the resolution. \# codevectors is 2048, \# of descriptors is 2048, codebook generated from CSIQ.}
	\begin{tabular}{lrrrrrrrr}
		\toprule
		\multicolumn{1}{c}{{resolution}} & \multicolumn{1}{c}{{ \# descriptors}} & \multicolumn{1}{c}{{ \# codevectors}} & \multicolumn{1}{c}{{plcc mean}} & \multicolumn{1}{c}{{plcc std}} & \multicolumn{1}{c}{{plcc median}} & \multicolumn{1}{c}{{srcc mean}} & \multicolumn{1}{c}{{srcc std}} & \multicolumn{1}{c}{{srcc median}} \\
		\midrule
		 $1920\times1080$ &   &   & 0.56  & 0.15  & 0.57  & 0.54  & 0.16  & 0.56 \\
		 $1280\times720$ & 64    & 2048  & 0.62  & 0.08  & 0.62  & 0.61  & 0.08  & 0.62 \\
		 $404\times720$ &  &  & 0.61  & 0.13  & 0.62  & 0.51  & 0.14  & 0.51 \\
		 \midrule
		 $1920\times1080$ &   &   & 0.60  & 0.14  & 0.64  & 0.58  & 0.16  & 0.63 \\
		 $1280\times720$ & 128   & 2048  & 0.63  & 0.08  & 0.64  & 0.62  & 0.08  & 0.63 \\
		 $404\times720$ &   &  & 0.65  & 0.13  & 0.65  & 0.56  & 0.14  & 0.59 \\
		 \midrule
		 $1920\times1080$ &  &  & 0.65  & 0.12  & 0.67  & 0.63  & 0.13  & 0.67 \\
		 $1280\times720$ & 256   & 2048  & 0.65  & 0.08  & 0.66  & 0.63  & 0.08  & 0.65 \\
		 $404\times720$ &  &   & 0.66  & 0.12  & 0.68  & 0.58  & 0.13  & 0.58 \\
		 \midrule
		 $1920\times1080$ &   &  & 0.70  & 0.11  & 0.73  & 0.68  & 0.12  & 0.71 \\
		 $1280\times720$ & 512   & 2048  & 0.68  & 0.07  & 0.68  & 0.65  & 0.08  & 0.67 \\
		 $404\times720$ &  &  & 0.70  & 0.11  & 0.70  & 0.60  & 0.13  & 0.60 \\
		 \midrule
		 $1920\times1080$ &  &  & 0.72  & 0.10  & 0.74  & 0.70  & 0.11  & 0.73 \\
		 $1280\times720$ & 1024  & 2048  & 0.69  & 0.07  & 0.69  & 0.66  & 0.07  & 0.66 \\
		 $404\times720$ & &  & 0.71  & 0.11  & 0.72  & 0.63  & 0.12  & 0.63 \\
		 \midrule
		 $1920\times1080$ &  &  & 0.72  & 0.10  & 0.74  & 0.70  & 0.11  & 0.72 \\
		 $1280\times720$ & 2048  & 2048  & 0.70  & 0.07  & 0.71  & 0.67  & 0.07  & 0.68 \\
		 $404\times720$ &  &  & 0.71  & 0.11  & 0.72  & 0.63  & 0.12  & 0.63 \\
		 \bottomrule
	\end{tabular}%
	\label{tab:LIVEVQC_resolution}%
\end{table*}%

\section{Discussion}
\label{sec:discussion}

The problem of synthetic image generation attarcted researches some time ago and continues to remain interesting with the advent of machine learning techniques as well as such quite recently emerged appoaches as GANS and diffusion models. In connection with this the goal of the present work was two-fold. First, we were interested in utilizing an apporach for image and video quality assessment which would be sufficiently effective in terms of correlations with MOS and, in the same time, built in such a way that to be close to real time algorithms in terms of compexity. We believe we were able to demonstrate the approach we applied satisfied both conditions. On the other hand, we investigated some properties of synthetic images built from simple primitives and applied these images as a set which can reflect some generic properties of sets of natural images. The rationale for that was the closeness of synthetic images generated with the model of this paper to real ones in terms of the scale-invariant power spectrum, related to second-order correaltions existing in natural images. We realize the approach and results are quite heuristic but still provide some evidence that our analysis and modifications of the codebooks algorithm we suggested gave rise to better results both in terms of performance and complexity reduction and also demonstrated potential relevance of synthetic data sets to structural description of sets of natural images.

One observations made in the work is that the role of codebooks in the algorithm has a few peculiarities. Specifically, a naive view may expect that the use of the data from a particular data set for constructing codevectors may result in better results if the model is trained on the same data set. This is not always the case: synthetic codebooks turn out to demonstrate better correlations not having any obvious relation to the train or test datasets. On the other hand, it is not possible to say that specific traits of the data set do not play any role when construction the model. Overall, in terms of the synthetic data used for the codebooks construction we demonstrate that it can be effective even when using some kind of principal component transform (e.g., whitening). This procedure is believed to correspond to some transformations occurring in the human brain when processing an image in the human visual system. However, we demonstrate that if we remove the whitening procedure, then the results of the method in terms of the algorithm can be sometimes superior to those obtaied with the application of ZCA. This may indicate that the properties of images not removed by whitening remain in connection with the properties of images.

\begin{table*}[htbp]
	\centering
	\caption{Performance comparison of IQA metrics on image datasets. The codebook model uses $2048$ descriptors and $2048$ codevectors. Stds were evaluated over $10$ splits in training/test sets. For each data set the codebooks model was trained using the same data set as well as the other models.}
	\begin{tabular}{crcccccccccccc}
		\toprule
               &                    & \multicolumn{4}{c}{ {CSIQ}} & \multicolumn{4}{c}{ {LIVE IQA}} & \multicolumn{4}{c}{ {TID2013}} \\ \cmidrule{3-14}
               &              Model & PLCC & STD  & SRCC &     STD      & PLCC & STD  & SRCC &       STD        & PLCC & STD  & SRCC &       STD       \\ \midrule
               &            BRISQUE & 0.74 &  -   & 0.81 &      -       & {\bf 0.94} &  -   & {\bf 0.93} &        -         & 0.57 &  -   & 0.63 &        -        \\ 
               &            DIIVINE & 0.78 &  -   & 0.80 &      -       & 0.90 &  -   & 0.89 &        -         & 0.57 &  -   & 0.64 &        -        \\ 
               &               NIQE & 0.71 &  -   & 0.63 &      -       & 0.90 &  -   & 0.91 &        -         & 0.39 &  -   & 0.32 &        -        \\ 
               &               Ours & 0.79 & 0.07 & 0.75 &     0.06     & 0.92 & 0.02 & 0.92 &       0.02       & {\bf 0.69} & 0.06 & 0.62 &      0.06       \\
               & Ours (luma+chroma) & {\bf 0.85} & 0.06 & {\bf 0.82} &     0.06     &  0.93 & 0.01 & {\bf 0.93} &       0.01       &  -   &  -   &  -   &        -        \\ \bottomrule 
	\end{tabular}%
	\label{comparison}%
\end{table*}%

\begin{table*}[htbp]
	\centering
	\caption{Performance comparison of VQA metrics on video datasets. The codebook model uses $2048$ descriptors and the same number of codevectors. Stds were evaluated over $10$ splits in training/test sets. For each data set the codebooks model was trained using the same data set as well as the other models.}
	\begin{tabular}{crcccccccc}
		\toprule
		                &                    & \multicolumn{4}{c}{ {LIVE VQC}} & \multicolumn{4}{c}{ {KONVID}} \\
		\cmidrule{3-10} &              Model & PLCC & STD  & SRCC &       STD        & PLCC & STD  & SRCC &      STD       \\ \midrule
		                &            BRISQUE & 0.64 & 0.06 & 0.59 &       0.07       & 0.66 & 0.03 & 0.66 &      0.04      \\ 
		                &          V-BLIINDS & 0.70 & 0.03 & 0.71 &       0.03       & 0.73 & 0.05 & 0.69 &      0.05      \\ 
		                &            VIDEVAL & 0.75 & 0.04 & 0.75 &       0.04       & 0.78 & 0.02 & 0.78 &      0.02      \\ 
		                &               Ours & 0.73 & 0.04 & 0.69 &       0.04       & 0.76 & 0.02 & 0.76 &      0.02      \\
		                & Ours (std pooling) & {\bf 0.76} & 0.03 & 0.73 &       0.04       & 0.77 & 0.02 & 0.77 &      0.03      \\ \bottomrule
	\end{tabular}%
	\label{comparison_video}%
\end{table*}%

\begin{table*}[htbp]
	\centering
\caption{Average correlations on image datasets for various types of codebooks. All experiments were carried out usin LIVE IQA dataset; the errors (standard deviations) were estimated as follows. $10$ splits (in proportion $8:2$) of the training dataset were made and the model was trained and tested $10$ times. After that std can be estimated. The numbers in bold show the best results for each combination of codevectors/descriptors number. The codebooks with ZCA in the names correspond to ZCA transform applied in the pipeline which those without this abbreviation did not use it. Noise means codebooks generated using random uncorrelated noise in each pixel. The synthetic codebooks were generated using (from the top to the bottom): 1) square and circle primitives, $\gamma=5$; 2) square and circle primitives, $\gamma=3$; 3) square, circle, and ellliptic primitives, $\gamma=5$.}
    \begin{tabular}{|lrrrrrr|}
    \toprule
    \multicolumn{1}{|c}{ codebook} & \multicolumn{1}{c}{ \# descriptors} & \multicolumn{1}{c|}{ \# codevectors} &\multicolumn{1}{c}{ PLCC} & \multicolumn{1}{c}{ std} & \multicolumn{1}{c}{ SRCC} & \multicolumn{1}{c}{ std} \\
    \midrule
     synthetic         &&&  {\bf 0.83} &  0.04 &   0.82 &  0.04  \\
     synthetic\_ZCA &&&  0.79 &  0.04 &   0.79 & 0.04  \\
     noise               &128&256&  0.76 &  0.04 &   0.75 &  0.05 \\
     noise\_ZCA       &&&  0.78 & 0.04 &   0.77 &  0.05 \\
     CSIQ                 &&&  0.81 &  0.03 &   0.80 &  0.03 \\
     CSIQ\_ZCA	&&&  {\bf 0.83} &  0.04 &   {\bf 0.83} &  0.04 \\
    \midrule
    synthetic          &&& 0.90  &   0.03 &    0.90 &   0.03 \\
     synthetic\_ZCA &&&   {\bf 0.92} &   0.02 &   {\bf 0.92} &   0.01 \\
     noise              &4096&256&   0.89 &   0.03 &    0.89 &   0.03 \\
     noise\_ZCA       &&&   0.89 &   0.03 &  0.89 &   0.03 \\
     CSIQ                 &&&   0.90 &   0.03 &   0.89 &   0.03 \\
     CSIQ\_ZCA         &&&   {\bf 0.92} &   0.02 &   {\bf 0.92} &   0.02 \\
    \midrule
     synthetic          &&&   0.93 &   0.02 &   0.93 &   0.02 \\
     synthetic\_ZCA &&&   0.91 &   0.03 &    0.91 &   0.03 \\
     noise               &4096&4096&   0.92 &   0.03 &   0.92 &   0.03 \\
     noise\_ZCA       &&&   0.92 &   0.02  &  0.92 &   0.02 \\
     CSIQ                 &&&   0.92 &   0.02 &    0.92 &   0.02 \\
     CSIQ\_ZCA        &&&   {\bf 0.94} &   0.01 &    {\bf 0.94} &   0.01  \\
    \bottomrule
    \end{tabular}%
  \label{tab:synthetic_codebooks}%
\end{table*}%


%



\section*{Acknowledgment}

The authors are grateful to their colleagues in Algorithm Innovation Lab for discussions.

\clearpage
\section*{Supplemental materials}
\label{sec:supplemental}

In this section we provide some supplementary information which can be of interest and demonstrates some additions to the main text, especially in terms of figures and tables.

\begin{figure*}
\centering
\includegraphics[width=0.9\linewidth]{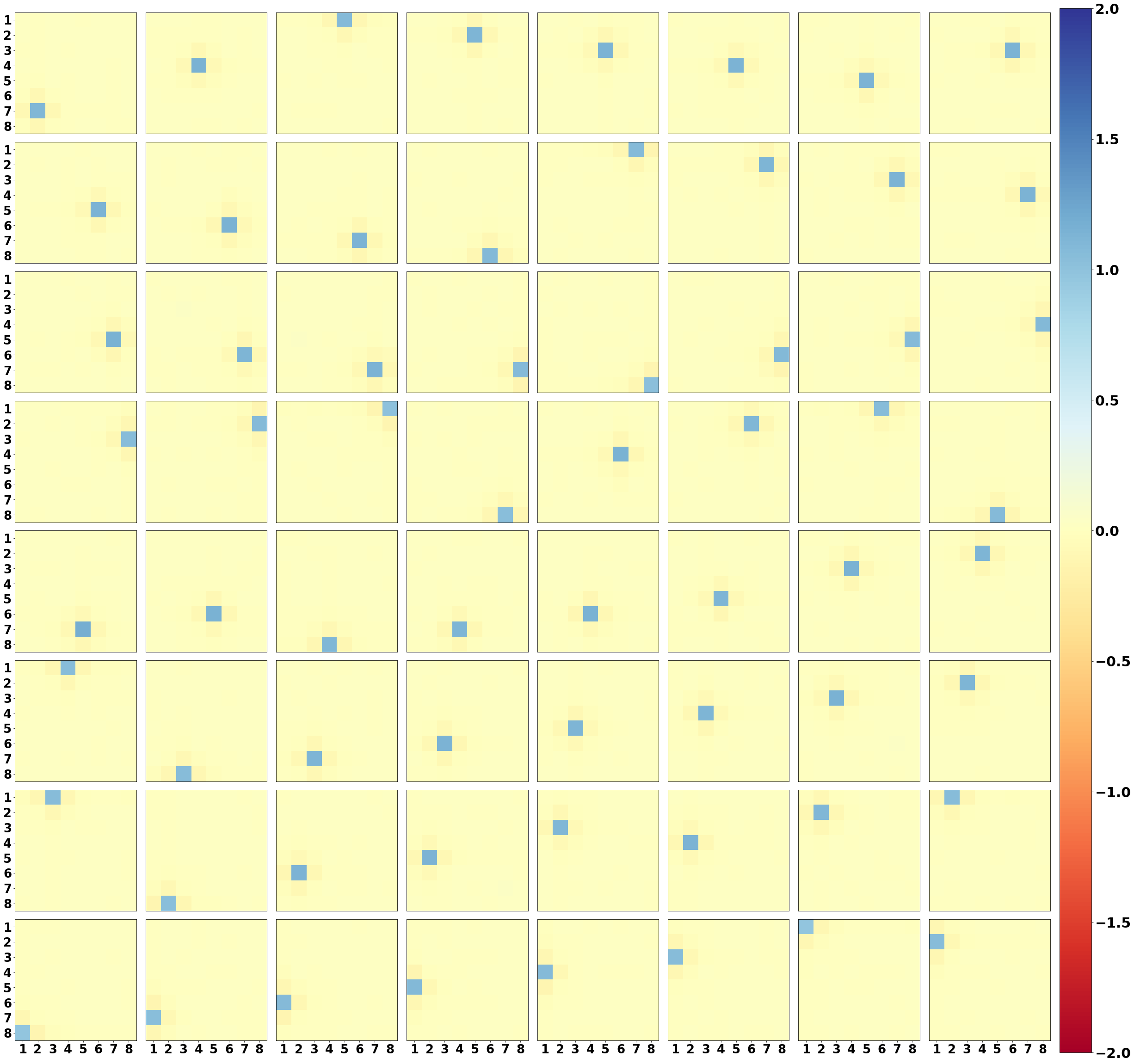}
\caption{Basis vectors of ZCA transform (columns of matrix $U$ in (\ref{ZCA})), located according to the decreasing eigen values (diagonal elements of $D$ in (\ref{ZCA})) for synthetic images generated by the model\cite{Koroteev2021}. The total number of $100$ synthetic images were generated and used for building codebooks. Each image was processed as explained in section \ref{sec:model} of the main text. Compared to fig. \ref{fig:zca_basis} the amplitudes are much more reduced which is accounted for by the smaller variations presented in the synthetic images both in terms of luminance and edges variability: the image contains very simple objects (fig. \ref{fig:synthetic_image}) in comparison with natural images.}
\label{fig:zca_basis_synthetic}
\end{figure*}

\newpage

\begin{table*}[htbp]
	\centering
	\caption{Average correlations on image datasets. The model trained on LIVE IQA dataset. Codebooks generated from CSIQ dataset.}
	\begin{tabular}{crcccccccccccc}
		\toprule
		&       & \multicolumn{4}{c}{ {CSIQ}}      & \multicolumn{4}{c}{ {LIVE IQA}}   & \multicolumn{4}{c}{ {TID2013}} \\
		\cmidrule{3-14}    \# descriptors & \# codevectors & PLCC  & STD   & SRCC  & STD   & PLCC  & STD   & SRCC  & STD   & PLCC  & STD   & SRCC  & STD \\
		\midrule
		
		\multirow{11}{*}{64} & 2     & 0.35  & 0.01  & 0.28  & 0.01  & 0.48  & 0.07  & 0.47  & 0.07  & 0.29  & 0.03  & 0.19  & 0.02 \\
						     & 4     & 0.44  & 0.01  & 0.36  & 0.01  & 0.67  & 0.05  & 0.66  & 0.05  & 0.34  & 0.01  & 0.25  & 0.02 \\
						     & 8     & 0.61  & 0.01  & 0.46  & 0.01  & 0.75  & 0.04  & 0.76  & 0.04  & 0.43  & 0.02  & 0.30  & 0.02 \\
						     & 16    & 0.62  & 0.01  & 0.47  & 0.01  & 0.80  & 0.04  & 0.80  & 0.04  & 0.37  & 0.02  & 0.28  & 0.02 \\
						     & 32    & 0.59  & 0.00  & 0.44  & 0.00  & 0.83  & 0.03  & 0.83  & 0.03  & 0.38  & 0.02  & 0.28  & 0.01 \\
		
		 					 & 64    & 0.61  & 0.01  & 0.47  & 0.01  & 0.84  & 0.03  & 0.84  & 0.03  & 0.40  & 0.01  & 0.30  & 0.01 \\
						     & 128   & 0.61  & 0.01  & 0.47  & 0.01  & 0.84  & 0.04  & 0.84  & 0.04  & 0.43  & 0.01  & 0.32  & 0.01 \\
						     & 256   & 0.62  & 0.01  & 0.47  & 0.01  & 0.84  & 0.03  & 0.85  & 0.03  & 0.44  & 0.01  & 0.32  & 0.01 \\
						     & 512   & 0.63  & 0.01  & 0.48  & 0.01  & 0.85  & 0.03  & 0.85  & 0.03  & 0.45  & 0.01  & 0.33  & 0.01 \\
						     & 1024  & 0.63  & 0.01  & 0.47  & 0.01  & 0.86  & 0.04  & 0.85  & 0.04  & 0.46  & 0.01  & 0.33  & 0.01 \\
						     & 2048  & 0.64  & 0.01  & 0.47  & 0.01  & 0.86  & 0.04  & 0.86  & 0.04  & 0.46  & 0.01  & 0.33  & 0.01 \\
		\midrule
		
		\multirow{11}{*}{2048}    & 2     & 0.31  & 0.01  & 0.25  & 0.01  & 0.55  & 0.09  & 0.54  & 0.09  & 0.38  & 0.03  & 0.27  & 0.04 \\
								  & 4     & 0.52  & 0.02  & 0.40  & 0.01  & 0.65  & 0.07  & 0.63  & 0.07  & 0.40  & 0.04  & 0.27  & 0.03 \\
								  & 8     & 0.61  & 0.01  & 0.46  & 0.01  & 0.78  & 0.02  & 0.77  & 0.02  & 0.43  & 0.02  & 0.36  & 0.02 \\
								  & 16    & 0.65  & 0.01  & 0.49  & 0.01  & 0.82  & 0.04  & 0.82  & 0.04  & 0.41  & 0.02  & 0.34  & 0.02 \\
								  & 32    & 0.67  & 0.01  & 0.55  & 0.01  & 0.88  & 0.02  & 0.88  & 0.02  & 0.54  & 0.02  & 0.45  & 0.02 \\
		
		   						  & 64    & 0.68  & 0.01  & 0.56  & 0.02  & 0.89  & 0.02  & 0.89  & 0.02  & 0.54  & 0.02  & 0.46  & 0.02 \\
							      & 128   & 0.70  & 0.01  & 0.59  & 0.01  & 0.90  & 0.02  & 0.90  & 0.02  & 0.53  & 0.01  & 0.45  & 0.02 \\
		  						  & 256   & 0.70  & 0.01  & 0.59  & 0.01  & 0.92  & 0.02  & 0.91  & 0.02  & 0.57  & 0.01  & 0.49  & 0.01 \\
		  						  & 512   & 0.70  & 0.01  & 0.58  & 0.01  & 0.91  & 0.02  & 0.91  & 0.02  & 0.58  & 0.01  & 0.49  & 0.02 \\
		 						  & 1024  & 0.71  & 0.01  & 0.59  & 0.01  & 0.92  & 0.02  & 0.92  & 0.02  & 0.59  & 0.01  & 0.50  & 0.02 \\
		 						  & 2048  & 0.71  & 0.01  & 0.58  & 0.01  & 0.92  & 0.02  & 0.92  & 0.02  & 0.59  & 0.01  & 0.50  & 0.01 \\
		 \bottomrule
	\end{tabular}%
	\label{correlations_live_csiq}%
\end{table*}%

\ifCLASSOPTIONcaptionsoff
  \newpage
\fi



\clearpage
\bibliographystyle{IEEEtran}
\bibliography{bibtex/bib/IEEEabrv,bibtex/bib/IEEEexample}
%




%




\end{document}